\documentclass[journal, compsoc]{IEEEtran} 
\usepackage{graphicx}
\usepackage[utf8]{inputenc}
\usepackage{hyperref} 
\usepackage{amssymb} 
\usepackage{amsmath} 
\usepackage{caption}
\usepackage{subcaption}
\usepackage{natbib}
\usepackage{siunitx} 
\usepackage{tabularx}
\setcitestyle{authoryear}

\makeatletter
\def\@seccntformatinl#1{\csname the#1dis\endcsname\hskip 1em\relax}
\makeatother

\begin{document}
%
\title{Graph-based methods for analyzing orchard tree structure using noisy point cloud data}

\author{Fred~Westling*,
    James~Underwood,
    Mitch~Bryson
    \thanks{All authors are with the University of Sydney}
    \thanks{*Correspondence: f.westling@acfr.usyd.edu.au}
}

\maketitle

\begin{abstract}
    Digitisation of fruit trees using LiDAR enables analysis which can be used to better growing practices to improve yield.  Sophisticated analysis requires geometric and semantic understanding of the data, including the ability to discern individual trees as well as identifying leafy and structural matter.  Extraction of this information should be rapid, as should data capture, so that entire orchards can be processed, but existing methods for classification and segmentation rely on high-quality data or additional data sources like cameras.  We present a method for analysis of LiDAR data specifically for individual tree location, segmentation and matter classification, which can operate on low-quality data captured by handheld or mobile LiDAR.
    Our methods for tree location and segmentation improved on existing methods with an F1 score of 0.774 and a v-measure of 0.915 respectively, while trunk matter classification performed poorly in absolute terms with an average F1 score of 0.490 on real data, though consistently outperformed existing methods and displayed a significantly shorter runtime.
\end{abstract}


\IEEEpeerreviewmaketitle

\section{Introduction}
\label{sec:introduction}

Understanding tree growth is an important consideration for commercial orchard operators.
There are many ways to manually measure growth factors, including mobile Leaf Area Index (LAI) measuring devices presented by \cite{confalonieri2013development} and \cite{francone2014comparison} or ceptometer sensors which \cite{ibell2015preliminary} showed could be used to study tree productivity.
However, manual measurements are difficult to automate and can have prohibitive restrictions including time required to take measurements, a requirement to measure in many locations, or weather limitations (such as a need for clear sky).
As an alternative, reality capture can be used to get digital models of the trees which can then be analysed.
Electromagnetic digitisation methods such as those presented by \cite{arikapudi2015orchard} are highly accurate but difficult to implement in practice on an orchard scale.
Cameras like those applied by \cite{underwood2016mapping} are cheap, accessible and flexible, but cannot always reconstruct geometric data.
LiDAR technology is rapidly improving, and can be a quick and detailed method of reality capture which provides large masses of data and is easy to automate.
\cite{wu2018estimating} measures changes in Leaf Area and Leaf Area Density for various tree crops using terrestrial LiDAR, \cite{westling2018light} presented a method which performed detailed analysis of tree growth factors using low quality LiDAR, and \cite{wu2020suitability} shows excellent results for mapping structural metrics like crown volume using airborne LiDAR which scales easily.
Here, we explore three separate operations which can be performed on low-quality LiDAR scans of orchard trees to enable further analyses, namely trunk location, individual segmentation and matter classification.

Previous works in trunk location in an orchard environment are typically focused on mobile platform localisation and mapping, and involve the use of multiple sensors.
\cite{bargoti2015pipeline} locate trunks primarily in the point cloud space using Hough transforms (89.7\% accurate), and then reproject the detections into into the camera frame to improve the results (95.8\%).
\cite{shalal2015orchard} similarly fuse laser scanner and camera data and distinguish between tree and non-tree objects, using the laser scanner to detect edge points and the camera for colour verification (96.64\%).
\cite{chen2018multi} instead fuse camera and ultrasonic data and train an SVM classifier to localise their robot using detected trunks (98.96\%).
However, all of these methods are working in a limited context, with a platform travelling parallel to rows of trees and processing on a frame-by-frame basis.

\begin{figure*}[t]
    \centering
    \begin{subfigure}[t]{0.32\textwidth}
        \includegraphics[width=\textwidth]{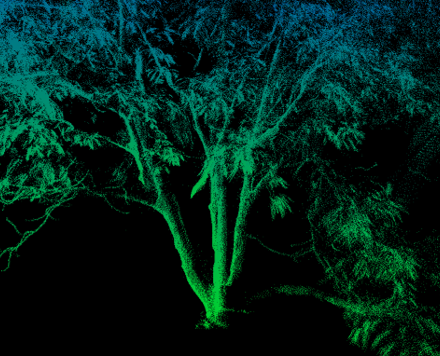}
        \caption{Tripod mounted Riegl VZ-400.  Capture time approximately 1 hour}
        \label{fig:quality:tripod}
    \end{subfigure}
    \begin{subfigure}[t]{0.32\textwidth}
        \includegraphics[width=\textwidth]{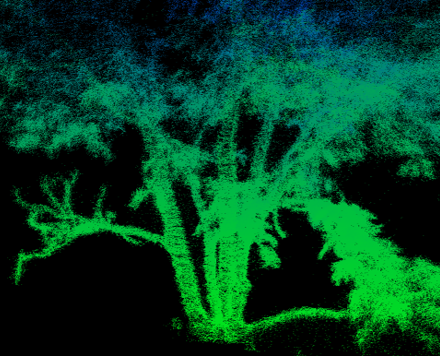}
        \caption{Handheld GeoSLAM Zeb-1.  Capture time approximately 5 minutes}
    \end{subfigure}
    \begin{subfigure}[t]{0.32\textwidth}
        \includegraphics[width=\textwidth]{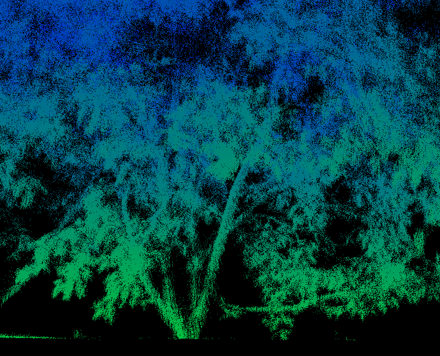}
        \caption{Mobile platform Velodyne HDL-64E.  Capture time approximately 10 seconds}
    \end{subfigure}
    \caption{LiDAR result quality comparison.  All three images represent the same tree, though the captures were at different times }
    \label{fig:quality}
\end{figure*}

Segmentation in this paper is defined as separating individual trees in the data, namely identifying which points belong to which trees.
This can allow better insights for end users, since results including tree growth parameters can be mapped to specific trees (\cite{underwood2016mapping}).
\cite{McFadyen2004} showed that yield improves with light interception and tree volume, but only up to a certain point, beyond which orchard crowding reduces yield over time.  If individual trees can be discerned, these effects can be better understood than if each row is just a wall of foliage.

Driven by the recent interest in autonomous driving applications, many of the current approaches to pointcloud semantic segmentation and classification operate on small pointclouds (e.g. \cite{guo2019deep}, up to 4096 points) as they are designed to run in realtime on single frames.
Most modern methods for segmentation in larger point clouds are in specific contexts with simple structures (\cite{poux2019voxel}) or work on simplified data from sampled CAD models rather than LiDAR data (\cite{liu2019densepoint}).
In agriculture specifically, a variety of methods have been explored.  \cite{underwood2016mapping} use cameras which have many advantages, but demonstrate difficulties in distinguishing overlapping branches, particularly since there is only one vantage point.
\cite{guan2015deep} uses euclidian distance clustering to segment trees, but the trees shown are spaced apart with minimal encroachment.
Good results were achieved by \cite{li2012new} with aerial LiDAR data using convex hulls, but this was a forestry application where again trees tend to be spaced out enough to make segmentation simple.
\cite{reiser2018iterative} presented good results on ground crops using very sparse point clouds, however their method relied on prior knowledge of crop spacing as well as a known location for each plant.
We aim to implement a method which works on very large point clouds with overlapping trees and no prior.

Classification in this paper is defined as assigning pointwise semantic meaning, specifically identifying which points represent leafy versus woody matter.
The key insight here is that woody matter (i.e. trunks and branches) are non photosynthetically active, and as explained by \cite{Ma2016a} there is benefit in measuring the amount of photosynthetically active material in a tree for growing purposes.
One application of this was presented by \cite{westling2018light}, who simulate the amount of light absorbed by trees digitized using LiDAR.  Identifying woody matter improves the quality of simulation with more accurate light transmission characteristics as well as better estimates of light absorption.

Trunk classification on pure point cloud data can be done in a wide variety of ways.
\cite{fritz2013uav} and others focused on tall trees with a single primary trunk apply cylinder fitting to detect that trunk, and classify surrounding points as leaves.
\cite{su2019extracting} uses a similar cylinder fitting method without the tall-tree assumption, but relies on high density scans containing minimal clutter points in order to identify cylindrical sections of point cloud.
A common approach to point cloud classification presented by several authors (e.g. \cite{Lalonde2006,Ma2016,brodu20123d}) involves using eigenvalue decomposition to describe patches of points into broadly three categories: planar, linear and random.  The patches can then be reliably classified as ground, trunk and leaf respectively, though this method is very sensitive to noise and can cause disconnected results due to its patch-based nature.
\cite{vicari2019leaf} presented an eigenvalue method which gets around this limitation by combining graph-based methods to integrate tree structure in the calculation.
\cite{Livny2010} similarly use a graph-based approach with optimised model fitting and generalised cylinders to reconstruct the skeletal structure of laser-scanned trees, while \cite{digumarti2018automatic} achieves good results in extracting the tree skeleton using local feature vectors.
Many of these methods rely on high quality data such as that captured by slow tripod-mounted scanners and are less effective on faster mobile data.

Static (tripod) LiDAR such as that used by \cite{vicari2019leaf}, \cite{Ma2016} and others produce excellent results as shown in Figure~\ref{fig:quality:tripod}.  However, use of static LiDAR requires time to set up and calibrate the position at each scan, and requires scanning from multiple positions for good coverage of each object, and the scans must then be combined to form a cohesive point cloud.  Due to these factors, they are not practical for scanning large areas like a commercial orchard setting.
At the other extreme, aerial LiDAR as used by \cite{windrim2018forest} can cover acres of land very rapidly, but the resultant data is much less accurate and much of it is occluded.  In particular, doing analyses below the top of the canopy becomes difficult.
Mobile LiDAR is a good compromise, allowing scanning of multiple acres per day with less occlusion.  However, the accuracy can suffer due to the limitations of necessary automated registration. \cite{makkonen2015applicability} found an RMSE of approximately 15-30mm using a handheld LiDAR, which is due to a combination of scanner accuracy, operator training and scanning procedure.  As shown in Figure~\ref{fig:quality}, the handheld and mobile options show features like leaves much less distinctly because of this.
Despite that, \cite{bauwens2016forest} showed that handheld LiDAR produces better coverage doing forest inventory than static LiDAR and \cite{ryding2015assessing} concluded that handheld sensors are efficient, cost effective and versatile for forest surveying.
Furthermore, LiDAR mounted on mobile platforms like that presented by \cite{underwood2016mapping} enables fully automated capture.
When the data can be captured and processed quickly, it can be applied to orchard-scale analysis, or analysis of individual trees over the entire orchard.
For these reasons, we are interested in developing methods which are applicable to low-quality point cloud data, and which ideally can be applied to data of variable quality.

Despite the lower quality, Mobile LiDAR has been used in a range of applications and industries, including building modelling in construction (\cite{sepasgozar2014implementation}), cultural heritage surveying (\cite{chan2016survey}) and mining (\cite{dewez2016handheld}).
As mentioned earlier, \cite{reiser2018iterative} was able to achieve good results doing ground crop plant segmentation with sparse mobile LiDAR data, but relied heavily on priors.
\cite{underwood2016mapping} used LiDAR on a mobile platform for orchard mapping and canopy volume.
\cite{westling2018light} presented a light environment simulation method using low-quality point clouds from handheld LiDAR.

Deep learning is an option for processing point clouds, though this presents its own challenges.
A review of the state of the art conducted by \cite{guo2019deep} found that most current approaches to point cloud object classification operate on point clouds up to 4096 points, which is insufficient for our analyses.
Methods using multi-view convolutional neural networks (e.g. \cite{su2015multi}) are unlikely to work in our context due to complex occlusions and varied environments.
Similarly, volumetric methods like that of \cite{wu20153d} or \cite{maturana2015voxnet} are similarly unsuited, since trees are large, varied in size, and highly complex.  These methods tend to be limited to voxels of size 32x32x32, which would lose a lot of detail in complex tree crops.
Direct point learning methods like PointNet (\cite{qi2017pointnet}) and its derivatives have mostly been used on standard datasets with perfect data sampled from 3D models, which produce far cleaner inputs than data from LiDAR.
\cite{guan2015deep} was able to use deep learning techniques to identify tree species by LiDAR, but not on the raw pointcloud, instead computing the waveform of the data and passing that into a neural net.
\cite{windrim2018forest} and \cite{xi2018filtering} use fully connected 3D CNNs to perform tree classification to good effect, though were applied to trees which are similar in size and shape and have little overlap.
\cite{kumar2019development} was able to identify that objects as trees or non-trees with 90\% accuracy, which is an operation on the macro scale and may not be applicable to small-scale features like branches and leaves.
Modern machine learning methods rely on extensive labelled datasets that are not readily available in orchard applications.  We instead focus on analytical methods rather than deep learning in order to avoid the need for labelled data in new contexts.

We present a system which, like \cite{vicari2019leaf}, uses graph-based methods to perform a range of tasks on point clouds in tree crops, with specific emphasis on handling low-quality and often overlapping data.
The method we present relies on the basic geometry of tree-like structures, namely that trees are connected by a network of woody matter, which is invariant to noise, fidelity and occlusion.

\section{Method}
\label{sec:method}

In this section, we first describe the methods used to collect or generate data for all experiments.  Then, the basic operation implemented is described, namely graph creation and search with an optional feature enrichment edge weighting scheme.  Finally, we describe the three operations to which the graph operation was applied.

The operations developed here were primarily using the ACFR Comma and Snark open-source libraries (\cite{acfrcomma}) and have been published to \url{https://github.com/fwestling/GraphTreeLS}.

\subsection{Data capture}

The data used in this study can be divided into real-world data captured using a LiDAR sensor from two orchards in Queensland, Australia and simulated (virtual) data designed to emulate the properties of LiDAR scans with generated tree object models.

\subsubsection{Scanning method}
\begin{figure}[ht]
    \centering
    \includegraphics[width=0.5\textwidth]{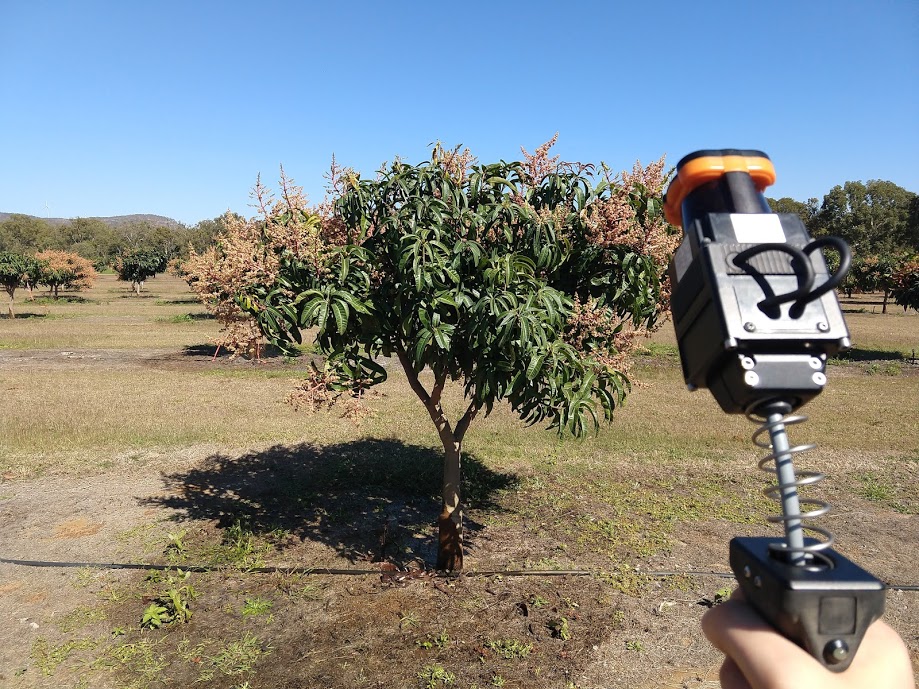}
    \caption{GeoSLAM Zeb1 sensor scanning a young mango tree}
    \label{fig:zeb}
\end{figure}

\begin{figure}[ht]
    \centering
    \begin{subfigure}[t]{\columnwidth}
        \centering
        \includegraphics[width=\textwidth]{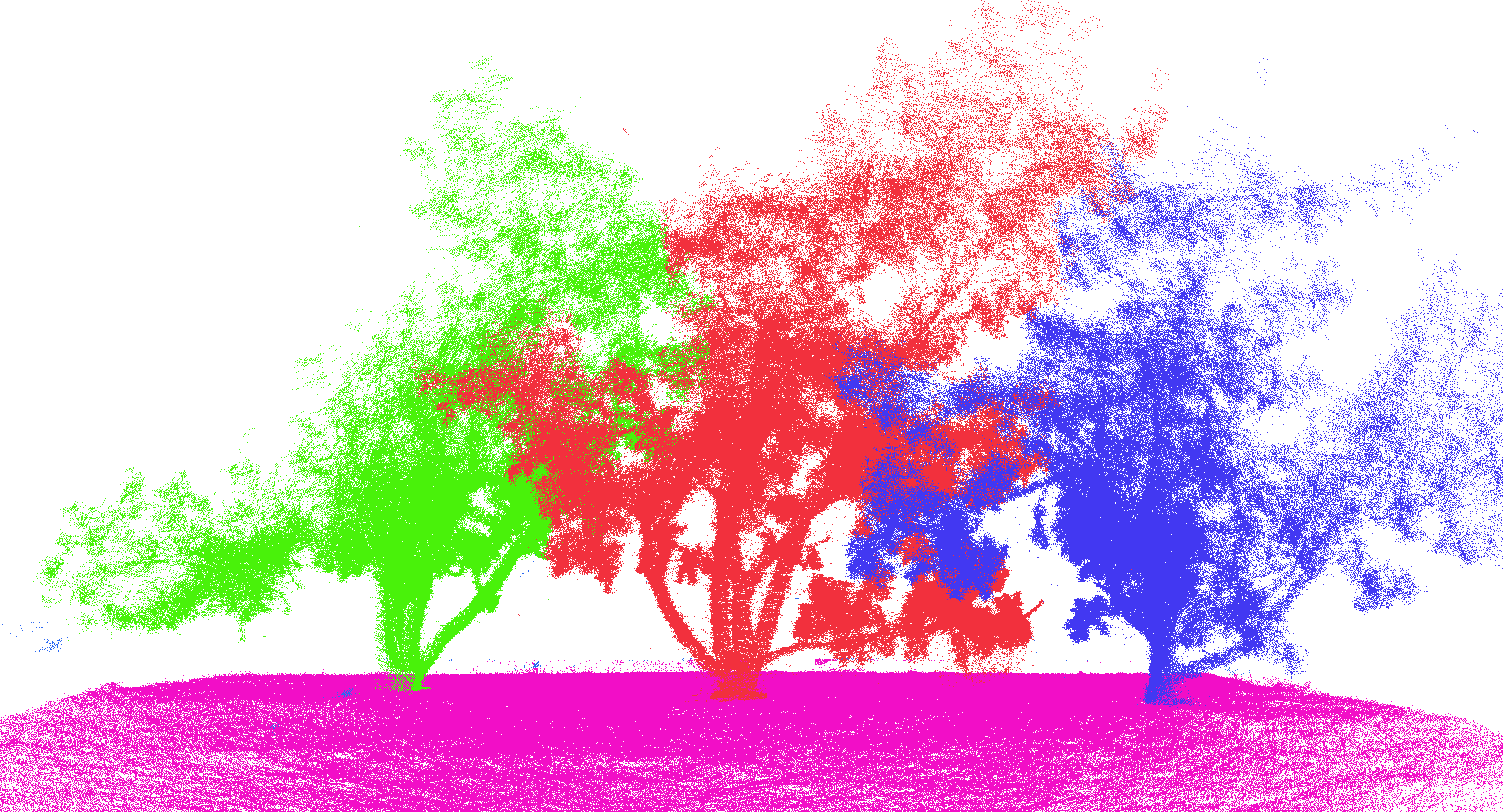}
        \caption{Individual tree segmentation}
    \end{subfigure}
    ~
    \begin{subfigure}[t]{\columnwidth}
        \centering
        \includegraphics[width=\textwidth]{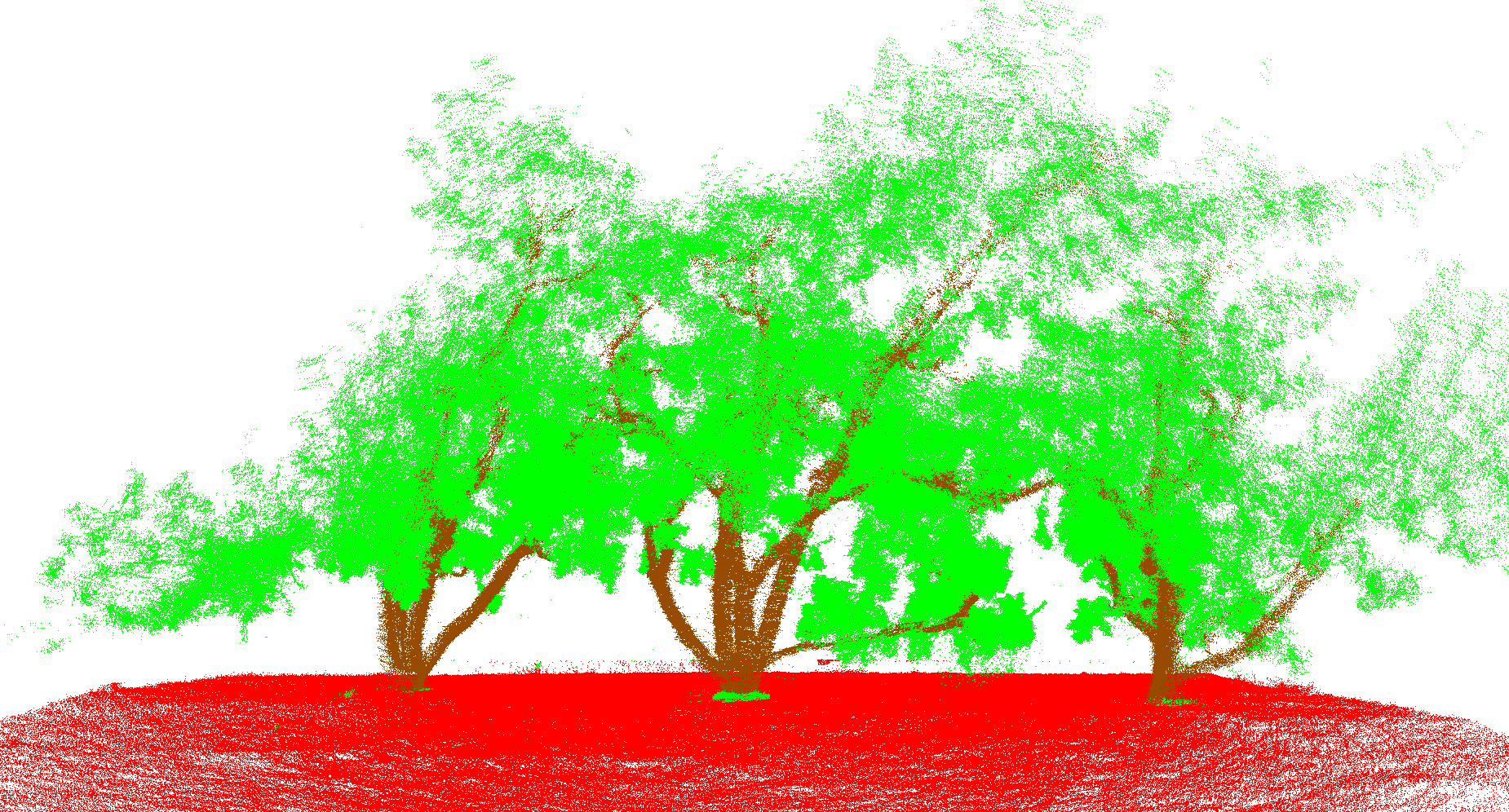}
        \caption{Matter classification}
    \end{subfigure}
    \caption{Stand of three hand-labelled avocado trees.  Viewing in colour is recommended.}
    \label{fig:method-label-real}
\end{figure}

The primary data used was captured at the Simpsons Farms commercial avocado orchard, with mature trees at 5m spacing with varied shape and often considerable overlap.
The trees were scanned using a handheld LiDAR, specifically the  GeoSLAM Zebedee 1 shown in Figure~\ref{fig:zeb}.  8 datum trees were selected to represent a variety of tree shapes, and each tree was scanned five times at different occasions over a period of two years during which they underwent fruit growth, harvest, and pruning.
The trajectory of the scan was kept reasonably consistent for each (accounting for operator error) and the path taken was designed to maximise coverage and minimise occlusion.
The focus of the scanning trajectory was on a single tree, so the surrounding neighbours were more sparsely sampled and less consistently covered.  To fix the quality at a consistent level, we cropped the point clouds for each datum tree down to just the tree and its two closest neighbours.
To provide a ground truth, we manually assigned labels to each point cloud in two steps.
First, we labelled the points as to which tree they belonged out of the three visible trees or the ground.
Second, we added a label to classify matter as leafy or woody matter.
This label was manually annotated using a 3D point cloud selection tool called "label-points", available in the open source "Snark" library (\cite{acfrcomma}).
Due to the nature of the scans, matter in the upper canopy was not clearly discernible, but large limbs were identified by eye.  To improve detection within the canopy, we split the point cloud into thin sections by height and annotated sections separately.
An example of one stand of avocado trees with both labels is presented in Figure~\ref{fig:method-label-real}.
For the avocado dataset, we generated 40 point clouds in this format, with 8 scans from 5 separate times.  Each point cloud contains 3 trees spaced 5 metres apart.
This dataset has been made public (\cite{westling2021avocado}).

A second set of real data was also collected, to allow for better testing of the trunk location operation.
This data was captured at a Queensland Government orchard intensification trial using young mango trees planted at three different densities (Low 8 x 6 m = 208 trees/ha, Medium 6 x 4 m = 418 trees/ha, High 4 x 2 m = 1250 trees/ha) are replicated and studied.
For these experiments, entire orchard blocks were scanned at a time to simulate the trajectory of a mobile sensor platform, so the data is less distinct for any individual tree but far more trees are included in each set.

\subsubsection{Virtual Dataset}
\begin{figure}[ht]
    \centering
    \includegraphics[width=\columnwidth]{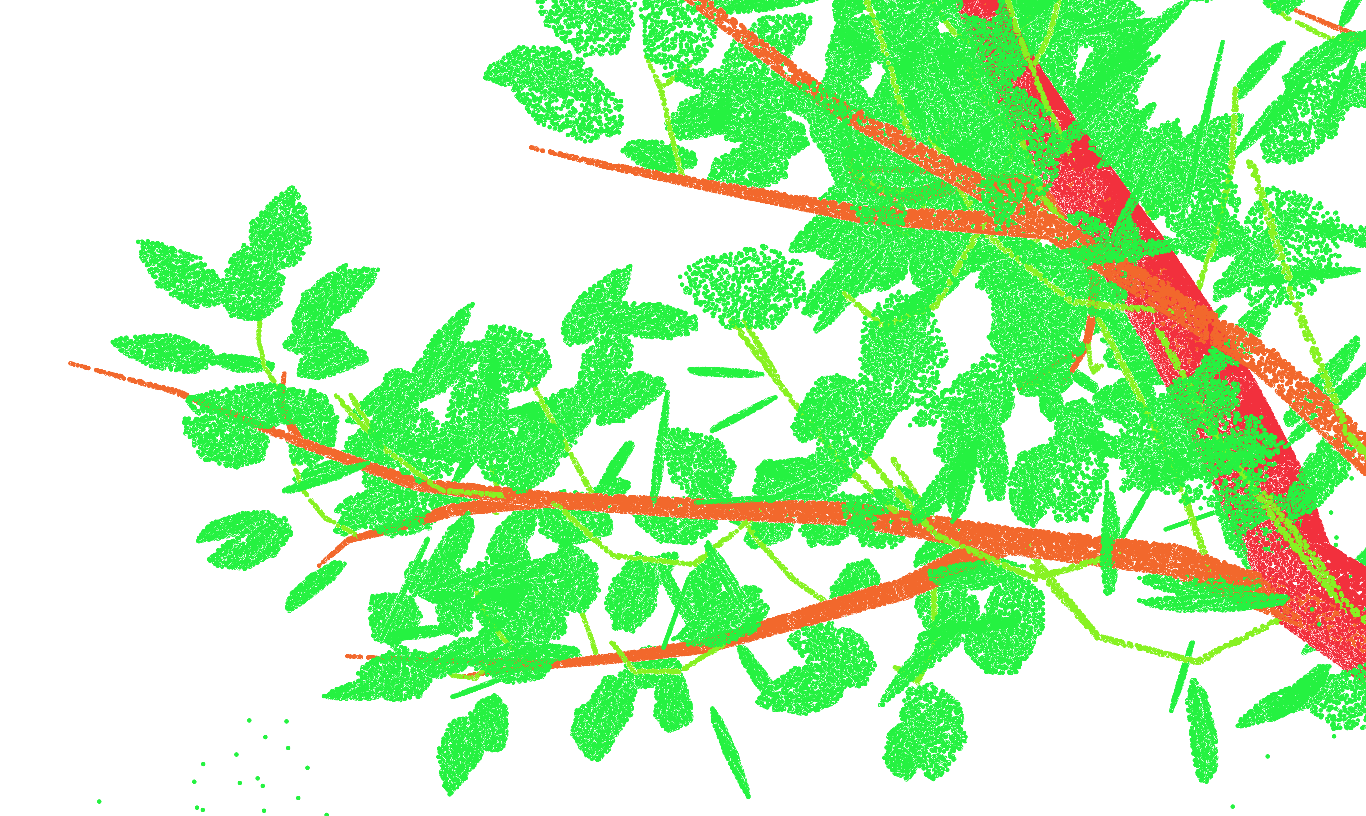}
    \caption{Virtual data with no noise, automatically labelled at three different levels of woody matter (red, orange, light-green) and leafy matter (green). Viewing in colour is recommended.}
    \label{fig:method-label-virtual}
\end{figure}

In the interest of more data for testing, as well as experimenting with different parameters, we produced a dataset of virtual tree scans using the SimTreeLS tool (\cite{westling2020simtreels}).
The scans generated by this tool were perfectly labelled, even in the difficult upper canopy, and are similar in quality to actual LiDAR scans, although the underlying trees only approximate the structure of real avocado trees.
With a non-deterministic and fully automated process, this dataset can be arbitrarily large and is perfectly labelled.
However, since simulated data is never a perfect substitute for real data, we use this as a supplemental set with which we can alter independent variables to better understand algorithmic robustness rather than as a primary indicator of quality.
An example of virtual generated data with no noise is shown in Figure~\ref{fig:method-label-virtual}, though the sets used in our experiments contain a small amount of gaussian noise, and is sampled according to the known LiDAR trajectories of the real data in order to provide a dataset with similar characteristics to the real data.

\subsection{Graph operation}
\begin{figure}[ht]
    \centering
    \begin{subfigure}[t]{\columnwidth}
        \centering
        \includegraphics[width=\textwidth]{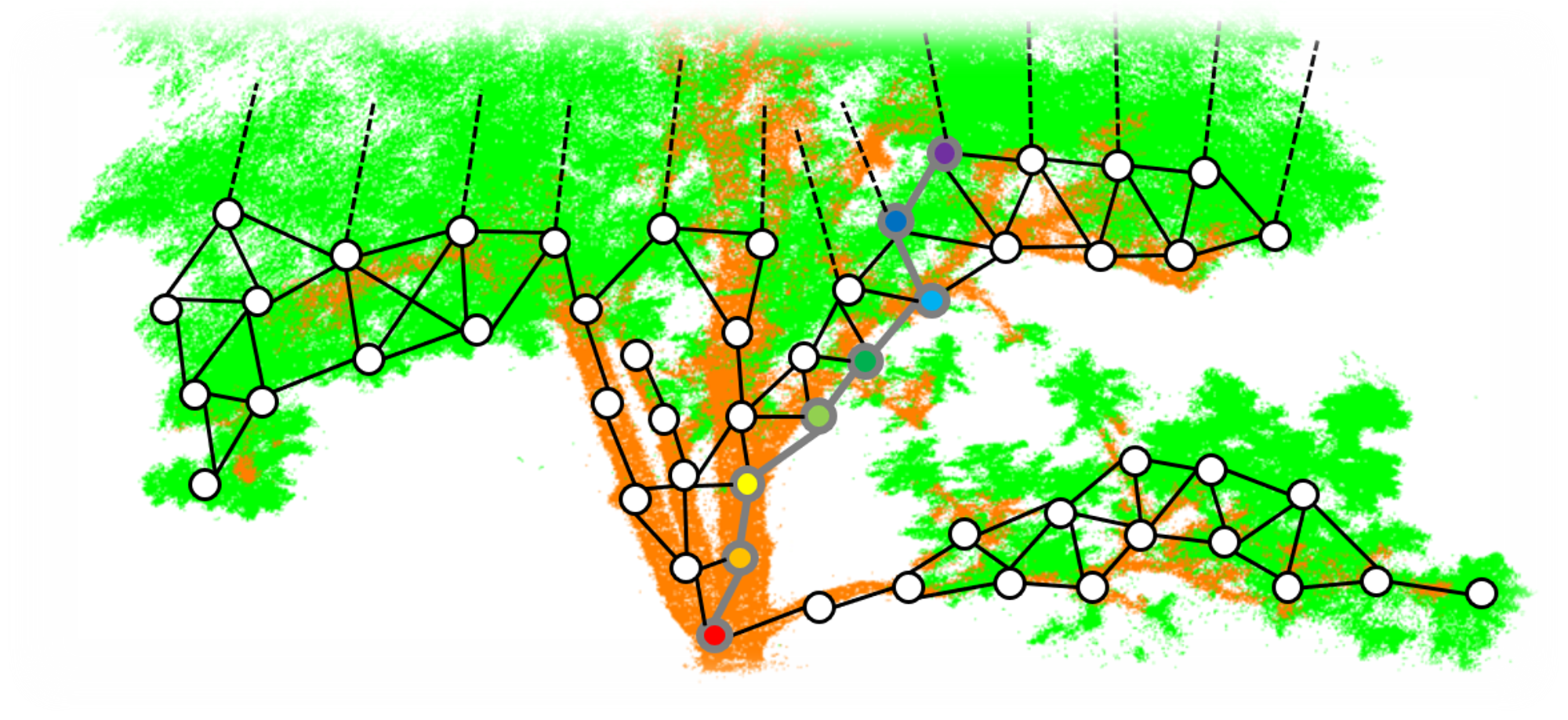}
        \caption{Graph with single path highlighted}
    \end{subfigure}
    ~
    \begin{subfigure}[t]{\columnwidth}
        \centering
        \includegraphics[width=\textwidth]{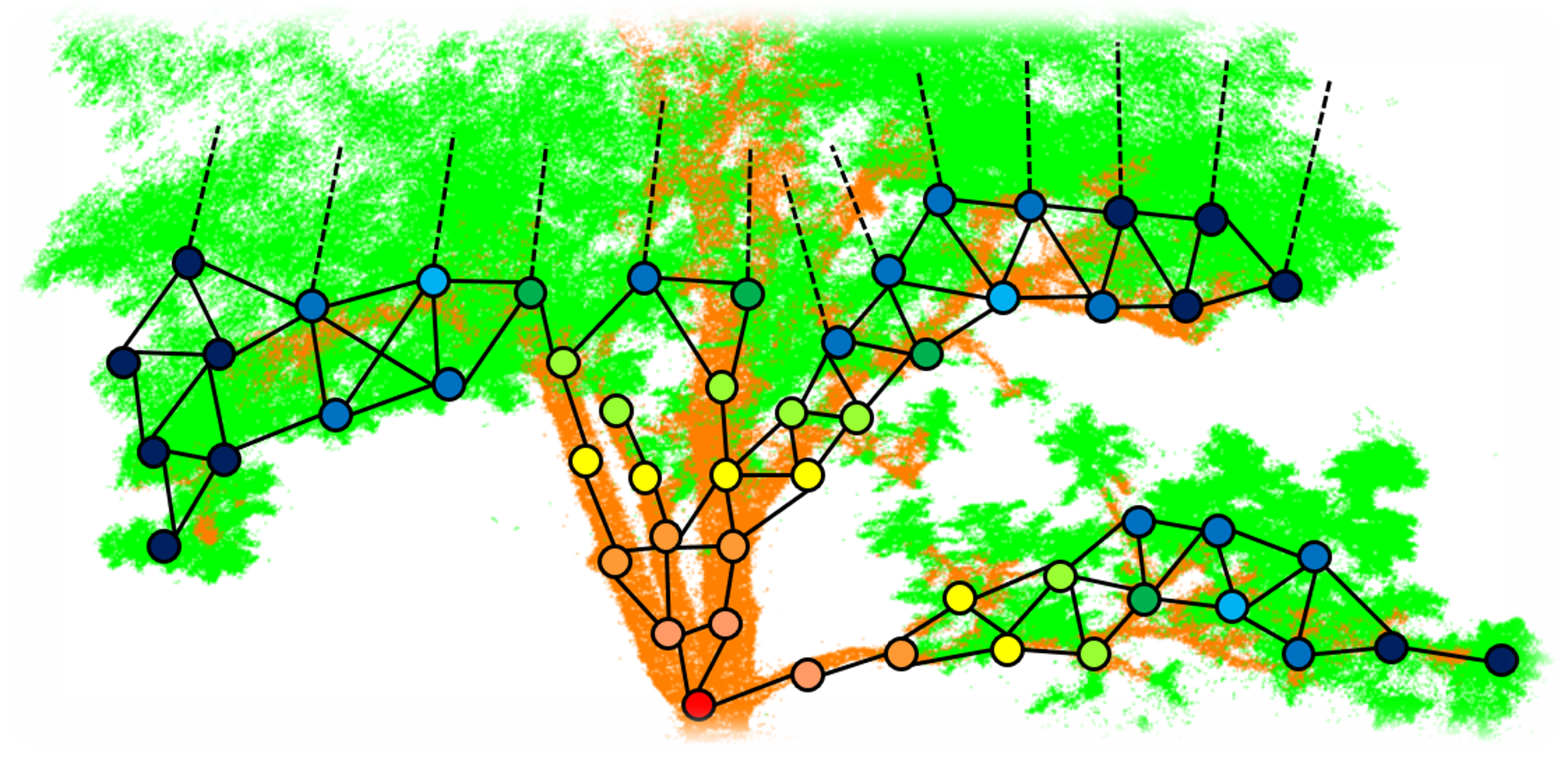}
        \caption{Graph with all paths aggregated}
    \end{subfigure}
    \caption{Illustration of the graph operation.  The point cloud is voxelised and each node is connected to its neighbours.  The shortest path is found from each node to the trunk node, and these paths are aggregated to produce a nodewise score.}
    \label{fig:method-graphop}
\end{figure}

The core method in our implementation involves a graph search over the point cloud, finding all paths through the cloud to any identified trunk points, and is illustrated in Figure~\ref{fig:method-graphop}.  Trunk points are defined as a single point per tree at the interface between the tree and the ground plane.
The ground is first removed by finding the local minima for each point in the Z axis within a lateral search radius ($R_g = 1m$).  Assuming sufficient LiDAR coverage such that any gaps in the scanned ground are smaller than $R_g$, this method can quickly identify the ground points and they can be excluded from graph construction since we know that no part of the ground can represent woody or leafy tree matter.
To normalize matter density and reduce search time, we first voxelise the point cloud at a given voxel size $v_s = 0.1m$ defined as the side length of a cubic voxel element.  The nodes of the graph are then defined as the average position of all the points in each voxel.
The edges for the graph are then defined.  In its simplest form, each node is connected to all neighbouring nodes withing a fixed radius $R_e = 0.15$.
When a query is performed, the shortest path is found using A* (\cite{hart1968formal}) from the trunk node of each tree represented in the point cloud to each node in the graph.
By aggregating all these paths, we can score each node according the number of times the node appeared in paths, which is proportional to the node's participation in the trunk network of the tree, and the length of the shortest path to that node.
We use these scores to achieve the various desired results.

\begin{figure}[ht]
    \centering
    \includegraphics[width=\columnwidth]{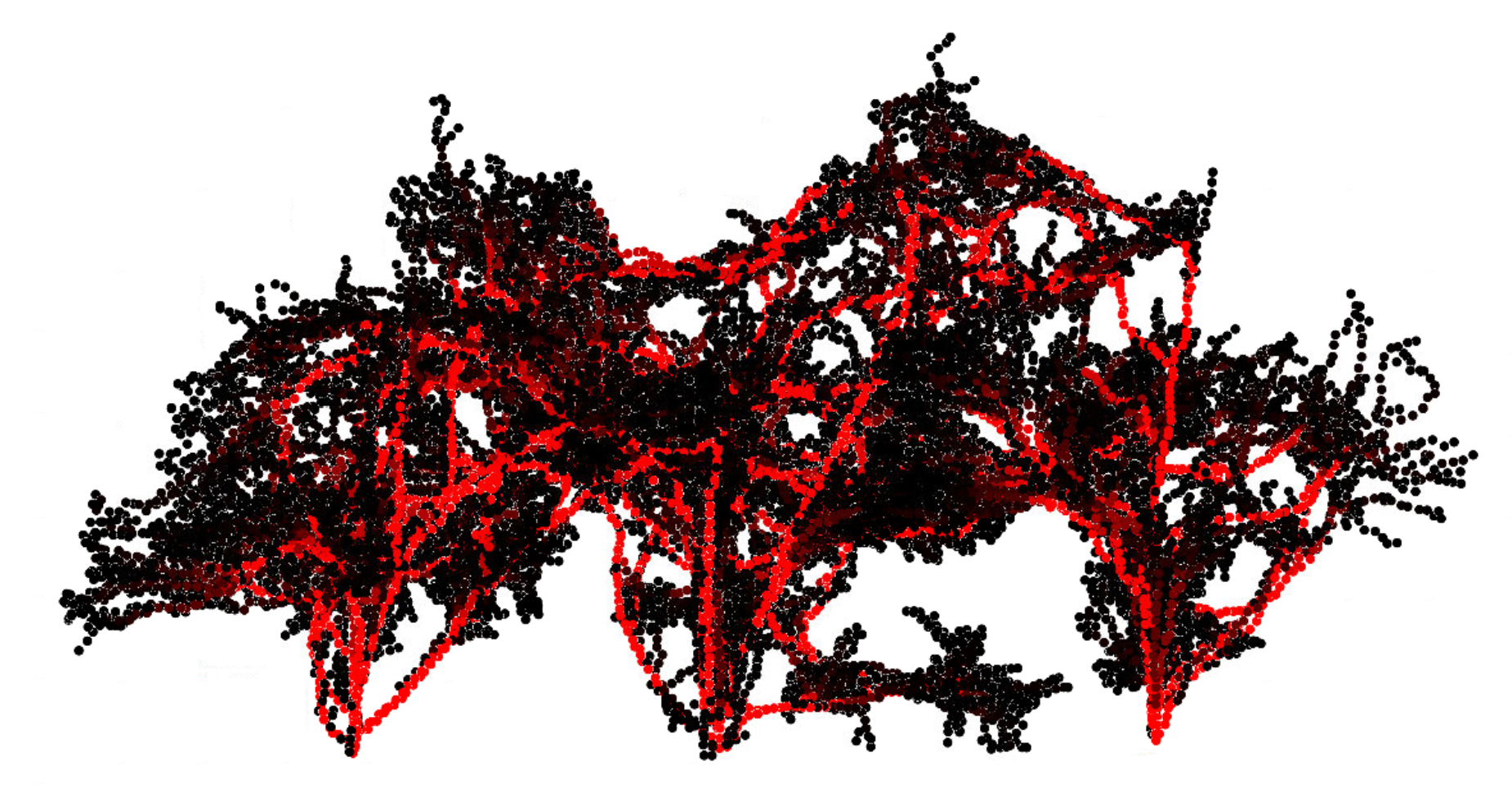}
    \caption{Visualisation of graph score in three overlapping avocado trees.  The score displayed is the number of paths in which each node appears, aggregated as the max per node.  Red points have a large score while black points have a low score, and nodes with a very small score have been removed for clarity. }
    \label{fig:method-graphscore}
\end{figure}

The reason this method is applicable is due to the basic geometry of trees.
The trunk and branches produce a network of connected matter to which all leaves are connected, and the paths through the canopy also form such a network, which overlaps in most cases.
Where paths exist through leaf matter, the point density tends to be very different, so edge weighting can be applied to encourage paths to traverse more trunk-like areas.
Figure~\ref{fig:method-graphscore} shows an example of how the graph score tends to follow woody matter.

\subsubsection{Enrichment}

\begin{figure}[ht]
    \centering
    \includegraphics[width=0.5\textwidth]{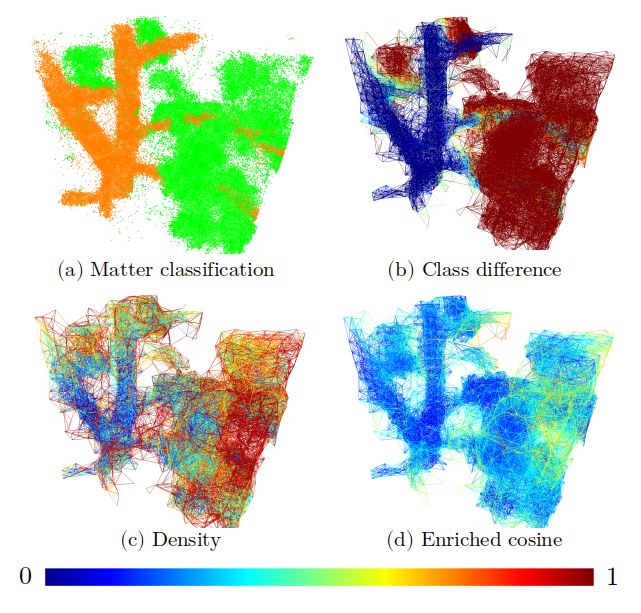}
    \caption{Sample of an avocado tree point cloud illustrating the graphing process.  Each subsampled point is connected to its neighbours, with edge weight coloured by different schemes. (a) shows the manually labelled classes of each point, with orange as trunk and green as leaf.  (b) has edge weights determined by difference in ground-truth class value, with trunk-trunk edges as 0 (blue) and leaf-leaf edges as 1 (red).  (c) has edge weights as the difference in voxel point count, penalising travel through low-density areas. (d) has edge weights as the cosine distance using all enriched features. In (c) and (d) red edges represent high-cost traversal while blue edges represent low-cost traversal.  Best viewed in colour.}
    \label{fig:method-graphop-minigraph}
\end{figure}

Multiple existing methods of point cloud classification utilise handcrafted features relying on local spatial features of the point cloud (\cite{Lalonde2006,Ma2016}).
However, the quality of the point clouds produced by handheld or otherwise mobile LiDAR is too low for methods relying solely on eigenvalue decomposition or local normals, since planar or cylindrical surfaces may present incorrectly due to occlusion, sensor noise, varying distance to the sensor, or movement due to wind.
That being said, computing these spatial features may enable better graph computation, so we implemented an enrichment approach which converts raw XYZ points into a set of additional features presented by \cite{poux2019voxel}, which are summarised in Table~\ref{tab:enrichment}.
During graph construction, we can use these features to introduce weights to the edges based on the relation between features of connected nodes.  Figure~\ref{fig:method-graphop-minigraph} visualises two such weighting schemes, namely difference in density and cosine similarity between all enriched features.

\begin{table}[ht]
    \begin{tabularx}{\columnwidth}{lX}
        {Feature}                                                          & {Description}                                                                                                                 \\
        \hline
        \\
        \multicolumn{2}{l}{{Eigen-based features}}                                                                                                                                                         \\
        \\
        {$\lambda_1,\lambda_2,\lambda_3$}                                  & {Eigen values of $V_{i,j,k}$}                                                                                                 \\
        {$\overrightarrow{v_1},\overrightarrow{v_2},\overrightarrow{v_3}$} & {Respective eigenvectors of $V_{i,j,k}$}                                                                                      \\
        {$\overrightarrow{v_3}$}                                           & {Normal vector of $V_{i,j,k}$}                                                                                                \\
        {$\lambda_a$}                                                      & {Anisotropy of $V_{i,j,k}$}                                                                                                   \\
        {$\lambda_e$}                                                      & {Eigen entropy of $V_{i,j,k}$}                                                                                                \\
        {$\lambda_l$}                                                      & {Linearity of $V_{i,j,k}$}                                                                                                    \\
        {$\lambda_o$}                                                      & {omnivariance of $V_{i,j,k}$}                                                                                                 \\
        {$\lambda_p$}                                                      & {Planarity of $V_{i,j,k}$}                                                                                                    \\
        {$\lambda_s$}                                                      & {Sphericity of $V_{i,j,k}$}                                                                                                   \\
        {$\lambda_v$}                                                      & {Surface variation of $V_{i,j,k}$}                                                                                            \\
        \\
        \multicolumn{2}{l}{{Geometric features}}                                                                                                                                                           \\
        \\
        {$\overline{V_{ix}},\overline{V_{iy}},\overline{V_{iz}}$}          & {Mean value of points in $V_{i,j,k}$ respectively along $\overrightarrow{e_x}$,$\overrightarrow{e_y}$,$\overrightarrow{e_z}$} \\
        {$\sigma_{ix}^2,\sigma_{iy}^2,\sigma_{iz}^2$}                      & {Variance of points in $V_{i,j,k}$}                                                                                           \\
        {$A_{V_p}$}                                                        & {Area of points in $V_{i,j,k}$ along $\overrightarrow{v_3}$}                                                                  \\
        {$A_V$}                                                            & {Area of points in $V_{i,j,k}$ along $\overrightarrow{e_z}$}                                                                  \\
        {$m$}                                                              & {Number of points in $V_{i,j,k}$}                                                                                             \\
        {$V_V$}                                                            & {Volume occupied by points in $V_{i,j,k}$}                                                                                    \\
        {$D_V$}                                                            & {Point density within $V_{i,j,k}$}                                                                                            \\
        \\
        \multicolumn{2}{l}{{Connectivity features}}                                                                                                                                                        \\
        \\
        {$C_H$}                                                            & {Number of horizontally adjacent voxels}                                                                                      \\
        {$C_V$}                                                            & {Number of vertically adjacent voxels}                                                                                        \\
        {$C_M$}                                                            & {Number of diagonally adjacent voxels}
    \end{tabularx}
    \caption{Enrichment features applied to each point cloud voxel $V_{i,j,k}$, as first presented by \cite{poux2019voxel}.  All features can be derived from just XYZ coordinates.}
    \label{tab:enrichment}
\end{table}

Multiple methods of computing edge weights from descriptors were investigated.
Ideally, the edge weighting scheme should present with a low weight for trunk to trunk edges, and a high weight for edges to or from leaf nodes.
The method chosen as that which best exhibited this behaviour was the cosine similarity of normalised values,

\begin{equation}
    S_C = \frac{\vec{f_A}\cdot \vec{f_B}}{\left \| \vec{f_A}\right \| \left \| \vec{f_B}\right \|}
\end{equation}

where $S_C$ is the cosine similarity between two graph nodes A and B while $\vec{f_A}$ and $\vec{f_B}$ are their respective feature vectors.

\subsection{Trunk detection}
\begin{figure}[ht]
    \centering
    \begin{subfigure}[t]{0.42\textwidth}
        \centering
        \includegraphics[width=\textwidth]{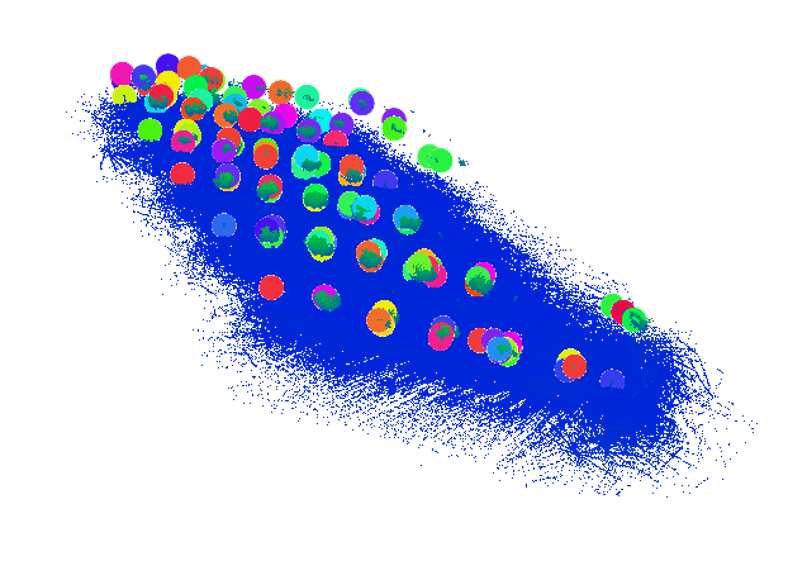}
        \caption{Point cloud with search targets}
    \end{subfigure}
    ~
    \begin{subfigure}[t]{0.42\textwidth}
        \centering
        \includegraphics[width=\textwidth]{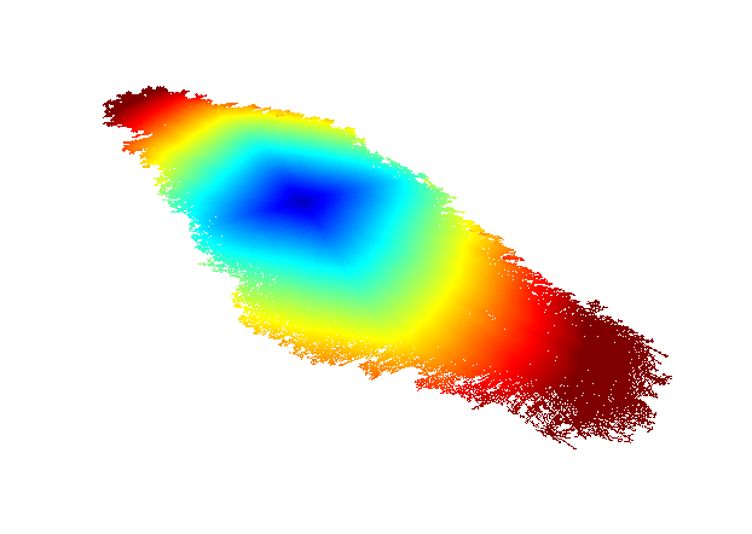}
        \caption{Graph score for single target}
    \end{subfigure}
    ~
    \begin{subfigure}[t]{0.42\textwidth}
        \centering
        \includegraphics[width=\textwidth]{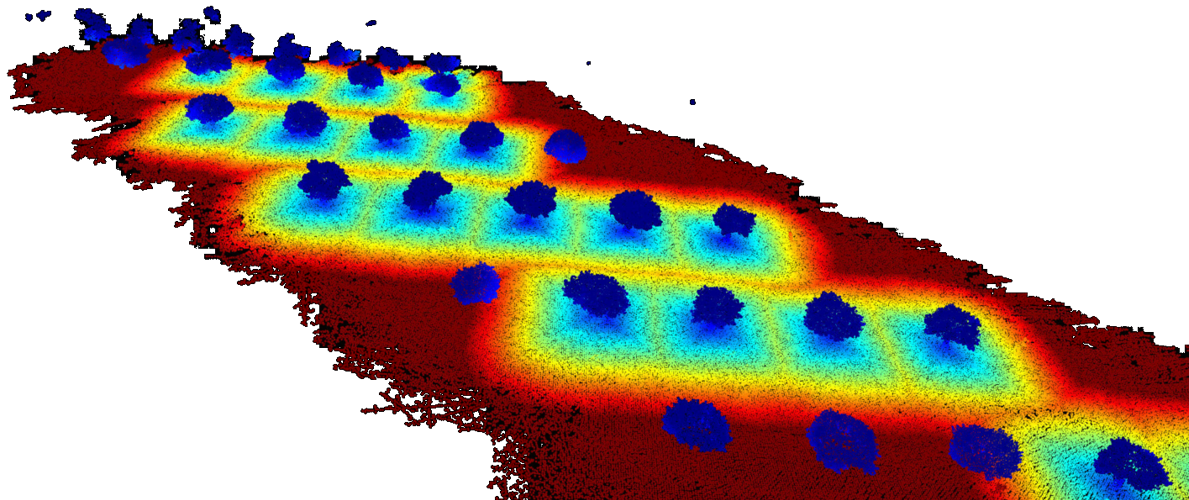}
        \caption{All graph searches aggregated by minimum score}
    \end{subfigure}
    \caption{Visualisation of the trunk finding operation.  First the search targets are generated on all non-ground points.  Then, the graph operation is used to score all ground points by path length and the result is aggregated to find ground entry points which are designated as trunks.}
    \label{fig:method-trunkfind}
\end{figure}

To use our graph operation method, we must know the location of one node per tree to serve as the start point for each path.
The trunk is a good choice for this node, since it is perfectly unambiguous as to which tree it belongs to, is at a consistent height, and is easy to manually label.
However, applying this method generically, autonomously or at scale requires a method for automatically locating trunk points.  We applied our graph operation to achieve this, as illustrated in Figure~\ref{fig:method-trunkfind}.
The graph construction used here is slightly different to that described.
The voxelisation is performed using a larger $v_s = 4m$, then we take a spatial subsample of the non-ground nodes to generate a list of source nodes.
We apply the graph search from each source node to all ground nodes, score each node by the length of the path to the target, and aggregate by taking the minimum score for each node.
The resulting score map across the ground nodes is characterised by a cluster of low scores around each ground-tree interface, which can then easily be filtered to find local minima which are then classified as trunk points.
Where multiple distinct trunk points exist in close proximity due to the presence of multiple targets per tree, a distance threshold is applied to cluster them as a single node.

When testing this method, we used real data with manual labelling as well as virtual data with automatically generated trunk locations using known tree spacings.
We counted the true positive (TP), false positive (FP) and false negative (FN) rate using a distance threshold of 1 metre to determine whether generated trunks matched the ground truth, and within the TP detections we reported the distance to the ground truth trunk.
Note that the trunks in the scans were approximately 0.2-0.5m in diameter, and the ground truth locations were on one side (for manually labelled real data) or in the center (for generated virtual data).

\subsection{Individual segmentation}
\begin{figure}[ht]
    \centering
    \includegraphics[width=0.5\textwidth]{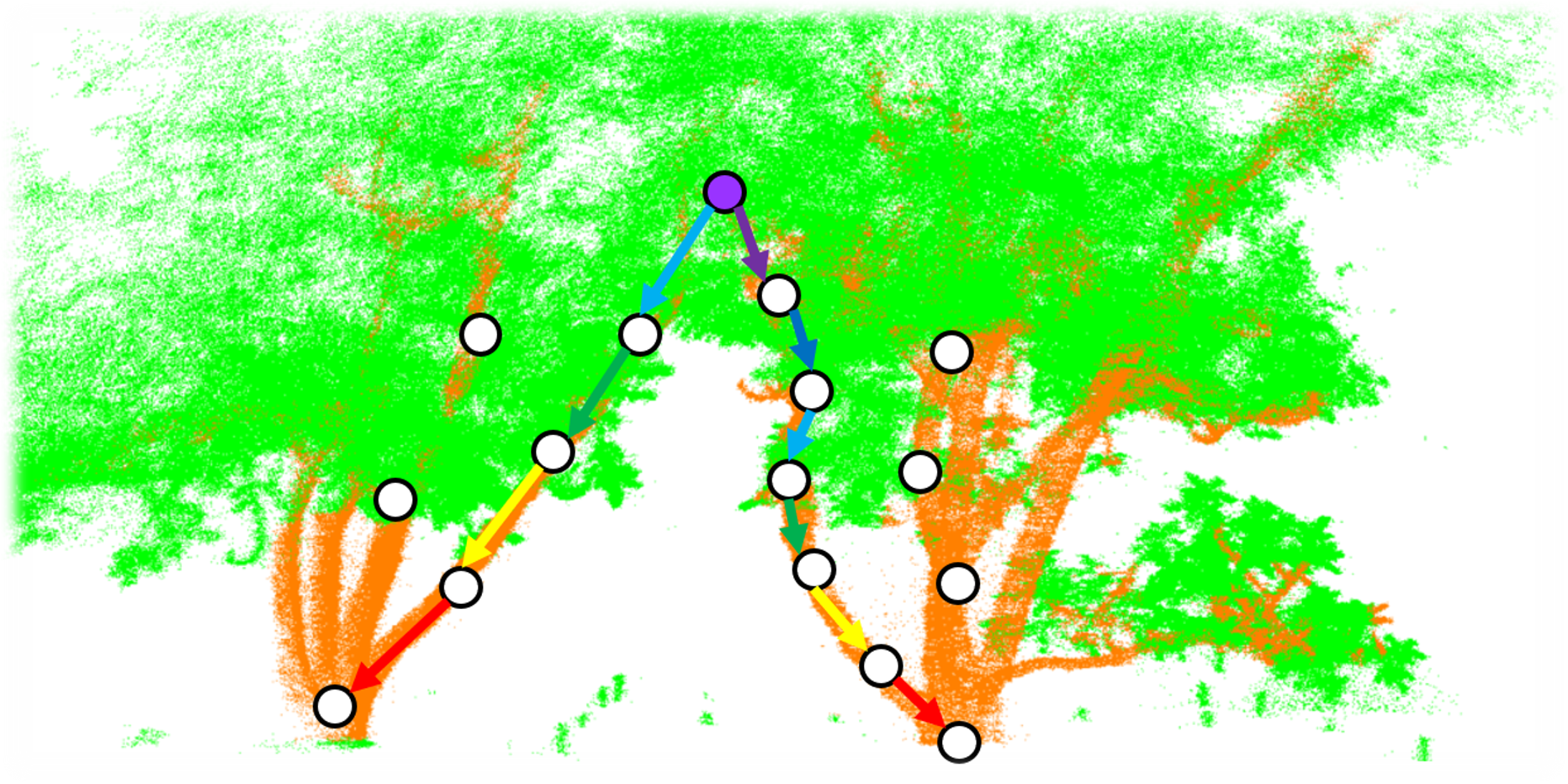}
    \caption{Illustration of the graph operation applied to tree segmentation.  When there is overlap, at least two paths to trunk will exist for each node, and the segmentation is determined by the shortest path.}
    \label{fig:method-graphop-seg}
\end{figure}

To perform tree segmentation using our graph operation, we track the paths from each trunk node to each node in the graph.
In cases where trees are sufficiently spaced that there are no non-ground paths between their trunk nodes, segmentation is equivalent to simple distance-based clustering.
For each tree node which was reached by a path, the node is allocated to the trunk node which originated that path.
When trees are close together or have long-reaching branches, they contain overlapping geometries and each tree node has multiple candidate trunk points.
In this case, we sort the candidates by the length of the shortest path from the node to the candidate trunk, assigning each node to its closest trunk.
This is visualised in Figure~\ref{fig:method-graphop-seg}.
Once each node, represented by a voxel, is assigned to a trunk, the segmentation is propagated to all points contained by the voxels.
Since the path length is a good representative for "real" (that is, connected through matter) distance from the trunk to each point, this produces better results than a simple distance-based segmentation which cannot handle overlapping canopies and branches.

\subsection{Matter Classification}
For matter classification, we aggregate the graph operation by the number of times each node appears in a path.
This per-voxel score, similar to the method presented by \cite{vicari2019leaf}, is computed as the ratio of logs over the entire graph:
\begin{equation}
    S_x = log(p_x) / log(p_M),
\end{equation}
where $S_x$ is the score for a particular node $x$, $p_x$ is the number of paths in which $x$ appears, and $p_M$ is the maximum value of $p$ over the whole graph.
The logarithm of path count is used since the range escalates quickly with the number of points, and by computing the ratio we normalise the score for different point clouds.  A point cloud with short trees which do not overlap will have a much lower $p_M$ than one containing tall trees with significant overlap, but $S_x$ compensates for this variability.

\begin{figure}[h]
    \centering
    \includegraphics[width=\columnwidth]{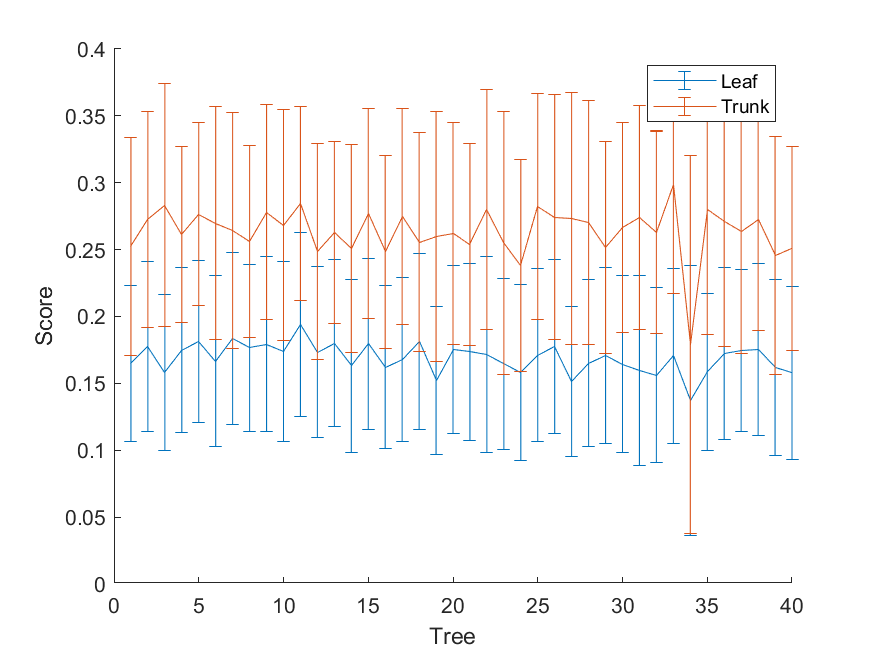}
    \caption{
        Mean and standard deviation of class score $S_x$ for both classes across our set of real data.
    }
    \label{fig:class-errorbar}
\end{figure}

Voxels are then classified with a threshold parameter, derived by comparing the scores of the classes as shown in Figure~\ref{fig:class-errorbar}.  Given the values observed, we selected a threshold of 0.216.

Finally, the voxelwise classification is propagated back to the component points by computing a large-neighbourhood average to "smooth" the classification and compensate for paths not passing through all nodes that make up a trunk.

\section{Results}
\label{sec:results}

\subsection{Trunk detection}

\begin{figure}[ht]
    \centering
    \includegraphics[width=0.5\textwidth]{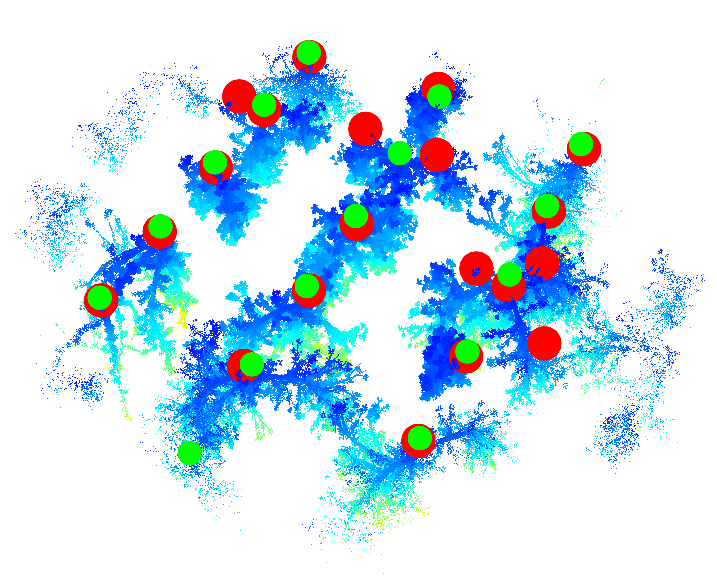}
    \caption{Example of trunk detection on one set of virtual data.  Trees are upside down and the ground has been removed for visual clarity. Red points are detected trunk locations, green points are ground truth.}
    \label{fig:results-ft-qual}
\end{figure}

\begin{figure}[ht]
    \centering
    \includegraphics[width=0.5\textwidth]{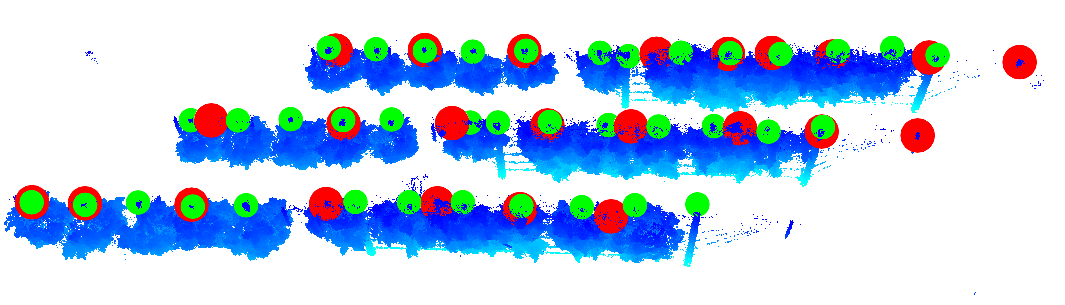}
    \caption{Example of trunk detection on one set of high density data.  Trees are upside down and the ground has been removed for visual clarity.   Red points are detected trunk locations, green points are ground truth.  Note the cases where closely packed trees are counted as one trunk.}
    \label{fig:results-ft-qual-hd}
\end{figure}

Trunk detection was applied to real data (stands of three avocado trees as well as entire blocks of mango trees) and virtual data, of which qualitative examples are presented in Figure~\ref{fig:results-ft-qual} (virtual trees) and Figure~\ref{fig:results-ft-qual-hd} (high density real trees).
The mango data is a mixture of low and medium density (not much overlap) and high density trees (significant overlap).
Generally trunk detection works well when trees are well defined, but not as well at the edges of scans where trees and the ground are poorly defined due to lower scanning density.
In virtual data, all generated trees are included in the results including these border trees.

Table~\ref{tab:results-findtrunks} display a summary of results on real and virtual data.
In both datasets, the majority of trunks were correctly detected, at an average distance of 0.357m and 0.452m respectively from the human labelled points.

\begin{table}[ht]
    \centering
    \begin{tabularx}{\columnwidth}{Xrrrr}
                                         & \multicolumn{2}{c}{Real} & \multicolumn{2}{c}{Virtual}                                                          \\
                                         & \multicolumn{1}{l}{Ours} & \multicolumn{1}{l}{Itakura} & \multicolumn{1}{l}{Ours} & \multicolumn{1}{l}{Itakura} \\
        Total TP                         & 874                      & 1700                        & 1095                     & 1344                        \\
        Total FP                         & 279                      & 2712                        & 774                      & 19911                       \\
        Total FN                         & 232                      & 218                         & 225                      & 221                         \\
        Precision                        & 0.758                    & 0.385                       & 0.586                    & 0.063                       \\
        Recall                           & 0.790                    & 0.886                       & 0.830                    & 0.859                       \\
        F1                               & 0.774                    & 0.537                       & 0.687                    & 0.118                       \\
        Average TP distance accuracy (m) & 0.357                    & 0.521                       & 0.452                    & 0.640
    \end{tabularx}
    \caption{Quantitative results from trunk detection experiments, aggregated over all sets of data used.  Results are presented for real and virtual data, for our method and that of \cite{itakura2018automatic}.  Distance error is the distance from the identified point to the manually selected trunk point, and is only reported for TP detections.}
    \label{tab:results-findtrunks}
\end{table}

\begin{figure}[ht]
    \centering
    \includegraphics[width=0.5\textwidth]{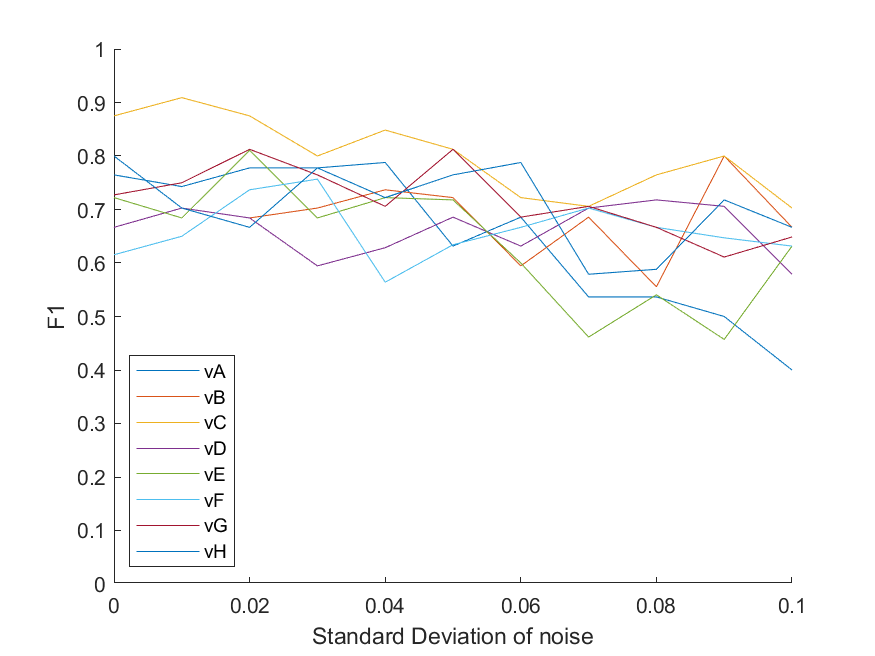}
    \caption{F1 score of trunk detection against artificial noise in virtual data.  Results are presented for 8 individual point clouds kept consistent in structure while noise is injected as gaussian noise with a standard deviation $\sigma$ ranging from 0 to 0.1 metres.}
    \label{fig:results-findtrunks-noise}
\end{figure}

Using virtual data, we intentionally inject different levels of noise when generating the point clouds.
Figure~\ref{fig:results-findtrunks-noise} shows the average F1 score across noise levels.

\subsection{Individual segmentation}

\begin{figure}[ht]
    \centering
    \begin{subfigure}[t]{\columnwidth}
        \centering
        \includegraphics[width=\textwidth]{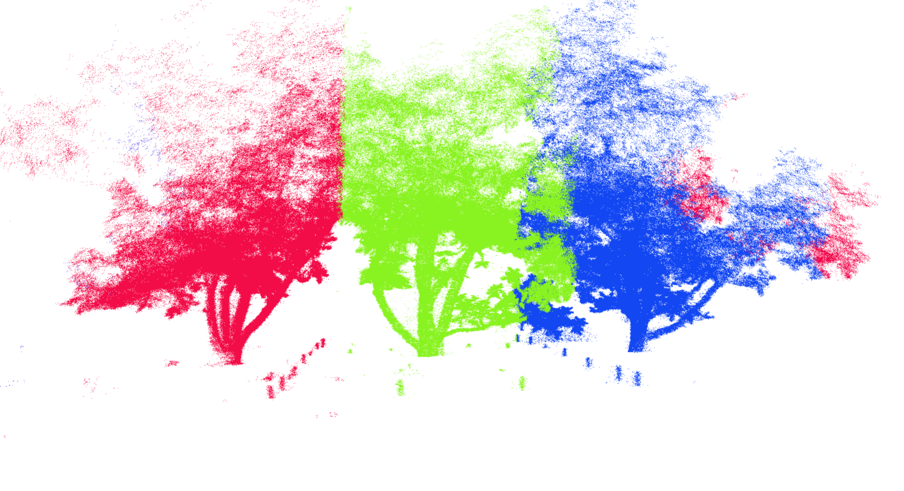}
        \caption{Closest-trunk method}
    \end{subfigure}
    ~
    \begin{subfigure}[t]{\columnwidth}
        \centering
        \includegraphics[width=\textwidth]{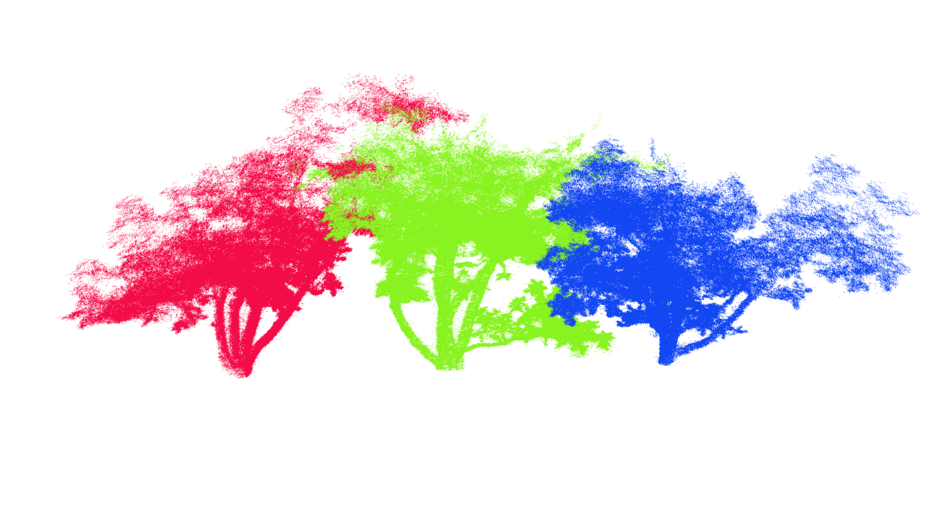}
        \caption{Graph method}
    \end{subfigure}
    \caption{Results from segmentation using closest-trunk method vs our method.  Isolated noise is discarded in the graph method as no path is it is not connected to the trunk, though this can easily be augmented by using the closest-trunk method for unclassified points.}
    \label{fig:results-seg-qual}
\end{figure}

For tree segmentation, we compare our method to a basic distance method where each point is allocated to the nearest trunk by straight-line distance.
Figure~\ref{fig:results-seg-qual} shows an example of a single avocado stand with both methods applied.
When scoring quantitatively, we use the v-measure cluster evaluation measure presented by \cite{rosenberg2007v}.  The v-measure provides a normalised score from 0 (no cluster overlap) to 1 (perfect overlap) with an arbitrary number of clusters without requiring labels to correspond.
Point clouds range in size from 3 trees to dozens per set.

The average v-measure for all real data was 0.915, while virtual data reported an average v-measure of 0.884.

\subsubsection{Virtual data}

\begin{figure}[ht]
    \centering
    \includegraphics[width=\columnwidth]{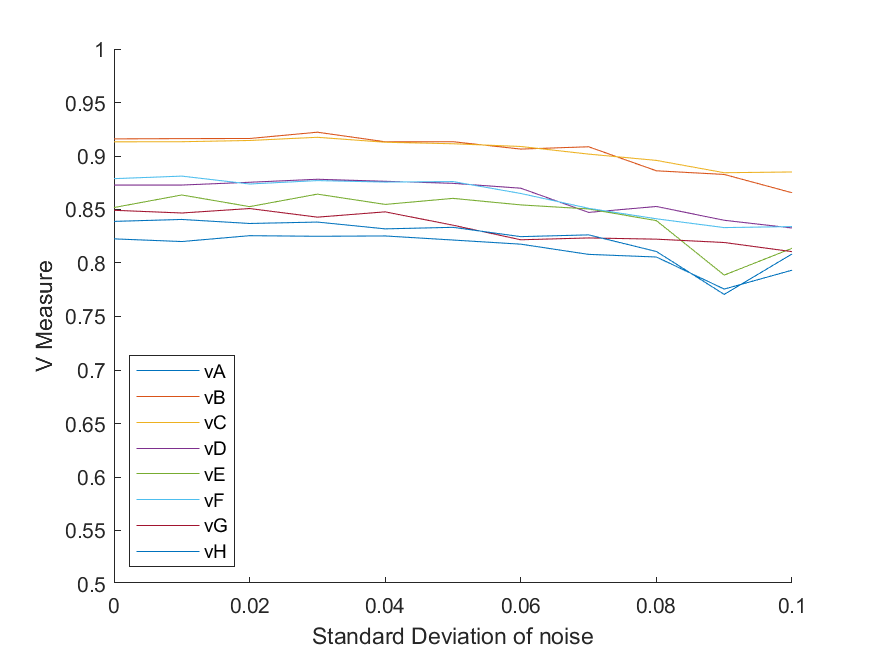}
    \caption{Segmentation results against virtual noise level.  8 randomly generated virtual stands are shown at different noise levels with a fixed tree spacing of 6m along the rows and 8m between rows.}
    \label{fig:results-seg-noise}
\end{figure}

\begin{figure}[ht]
    \centering
    \includegraphics[width=\columnwidth]{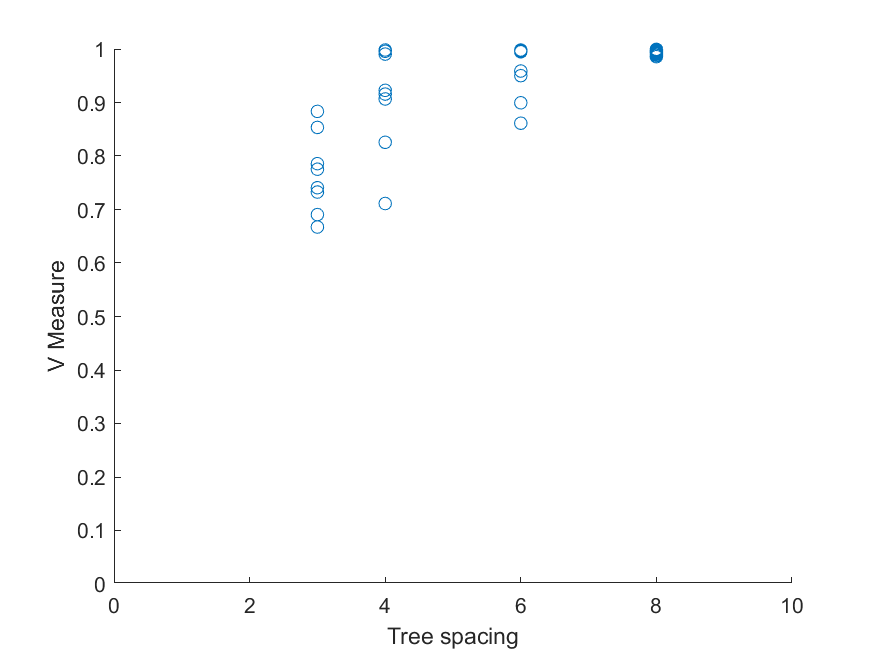}
    \caption{Segmentation results against virtual tree spacing.  5 randomly generated virtual stands are shown with different spacings, fixed at a noise level of 0.02 metres.}
    \label{fig:results-seg-spacing}
\end{figure}

Since we have more control of the dataset in the virtual space, we perform testing over noise levels, presented in Figure~\ref{fig:results-seg-noise}, and tree spacing, presented in Figure~\ref{fig:results-seg-spacing}.

\subsubsection{Real data}

\begin{figure}[ht]
    \centering
    \includegraphics[width=\columnwidth]{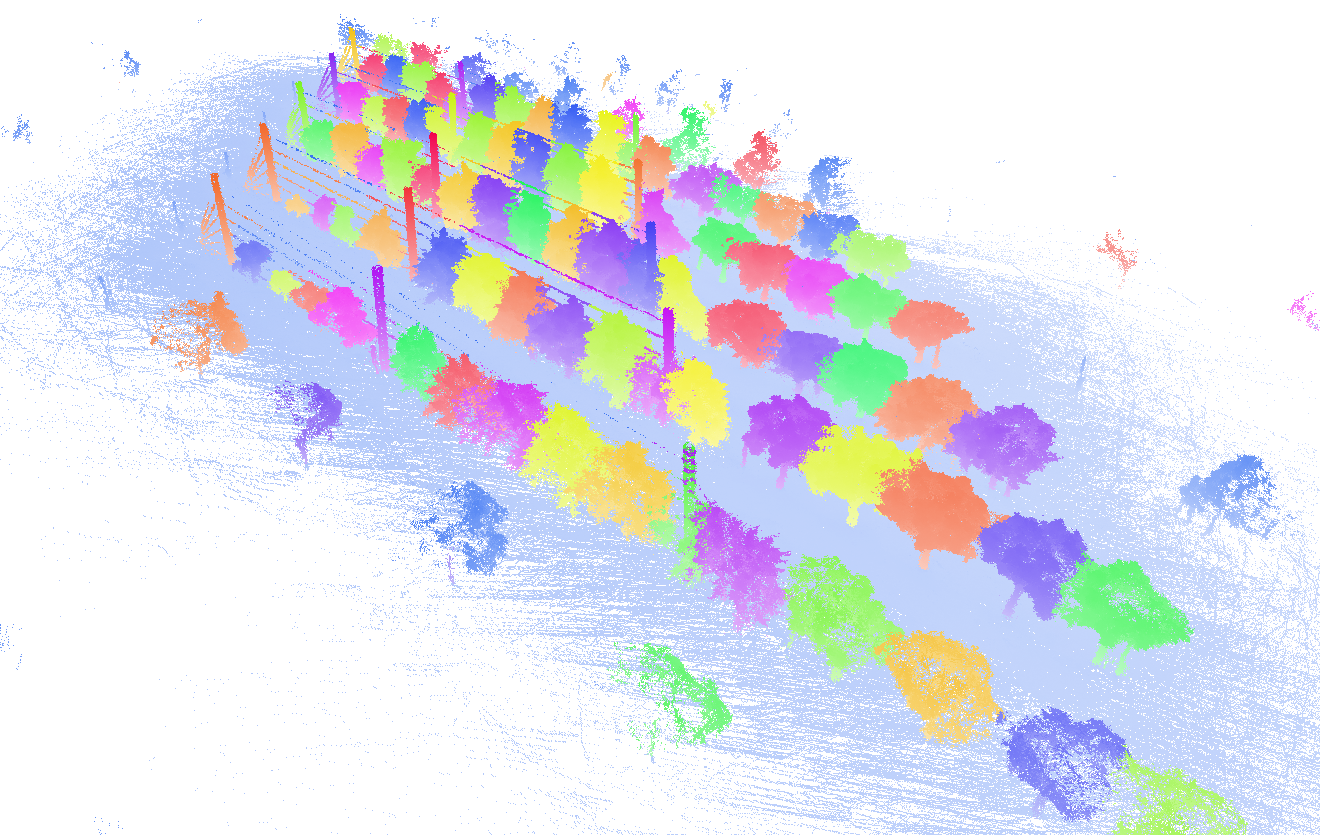}
    \caption{Automated segmentation results on high density mango trees with known trunk locations.}
    \label{fig:results-seg-qual-real}
\end{figure}

For comparison against real data, we focus on the avocado trees which are relatively complex in structure, closely spaced and display considerable overlap.
However, an example of our segmentation applied to high density mango trees is shown in Figure~\ref{fig:results-seg-qual-real}.

\begin{figure}[ht]
    \centering
    \includegraphics[width=\columnwidth]{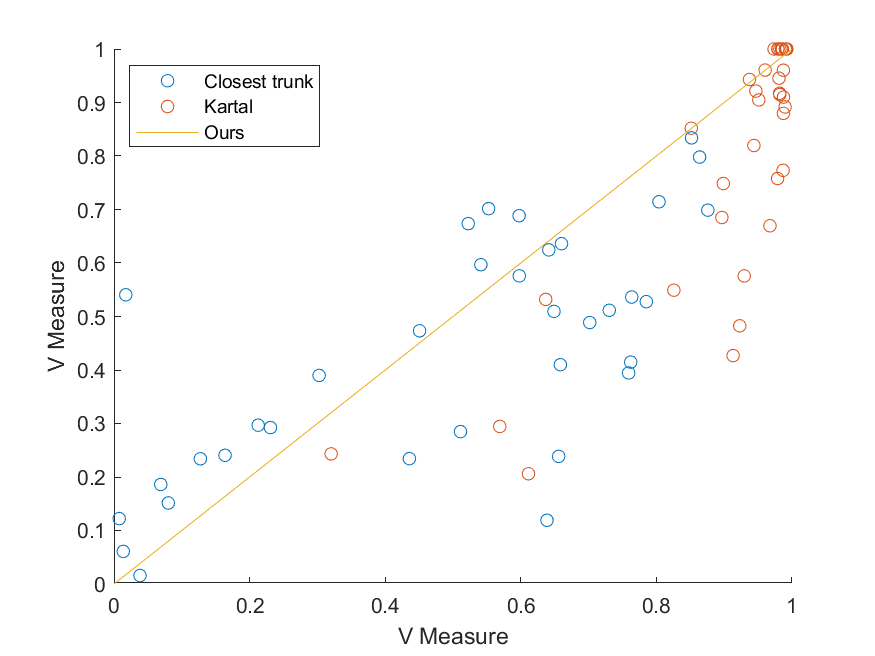}
    \caption{Results of our segmentation against the closest-trunk method and the KMeans method presented by \cite{kartal2020segmentation} on real data, excluding data where the two methods agree.  The yellow line represents parity, so points above this line perform better than our method.}
    \label{fig:results-seg-comp-real}, which shows the v-measure results between our method the others specifically only on points where both methods agree.
    The results show our method outperforming mostly that of Kartal for, but interestingly the closest-trunk method performs better lower values of v-measure.  This could be explained by
\end{figure}

\begin{figure}[ht]
    \centering
    \includegraphics[width=\columnwidth]{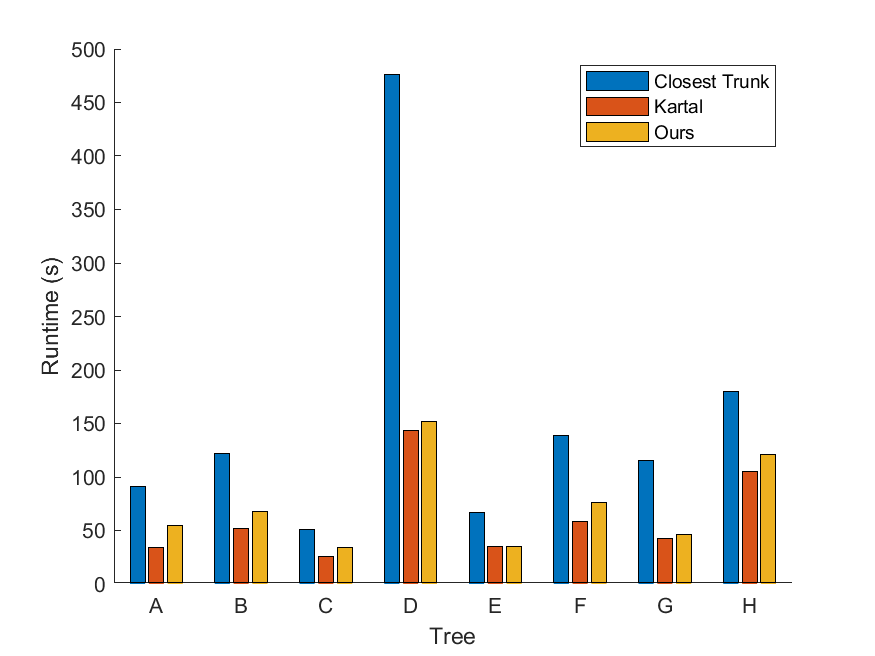}
    \caption{ Runtimes for segmentation on real data}
    \label{fig:results-seg-times-real}
\end{figure}

Meanwhile the quantitative results on the avocado data are presented in Figure~\ref{fig:results-seg-comp-real}, which shows the v-measure results between our method the others specifically only on points where both methods agree.
The results show that , our method outperforming mostly that of Kartal for, but interestingly the closest-trunk method performs better for lower values of v-measure.  This could be explained by  with runtimes in Figure~\ref{fig:results-seg-times-real}.
We display the results for applying our method with no edge weights, as well as applying it with the edges weighted as per the cosine similarity score between the enriched features of connected nodes.

\subsection{Matter Classification}

\begin{figure}[ht]
    \centering
    \begin{subfigure}[t]{\columnwidth}
        \centering
        \includegraphics[width=\textwidth]{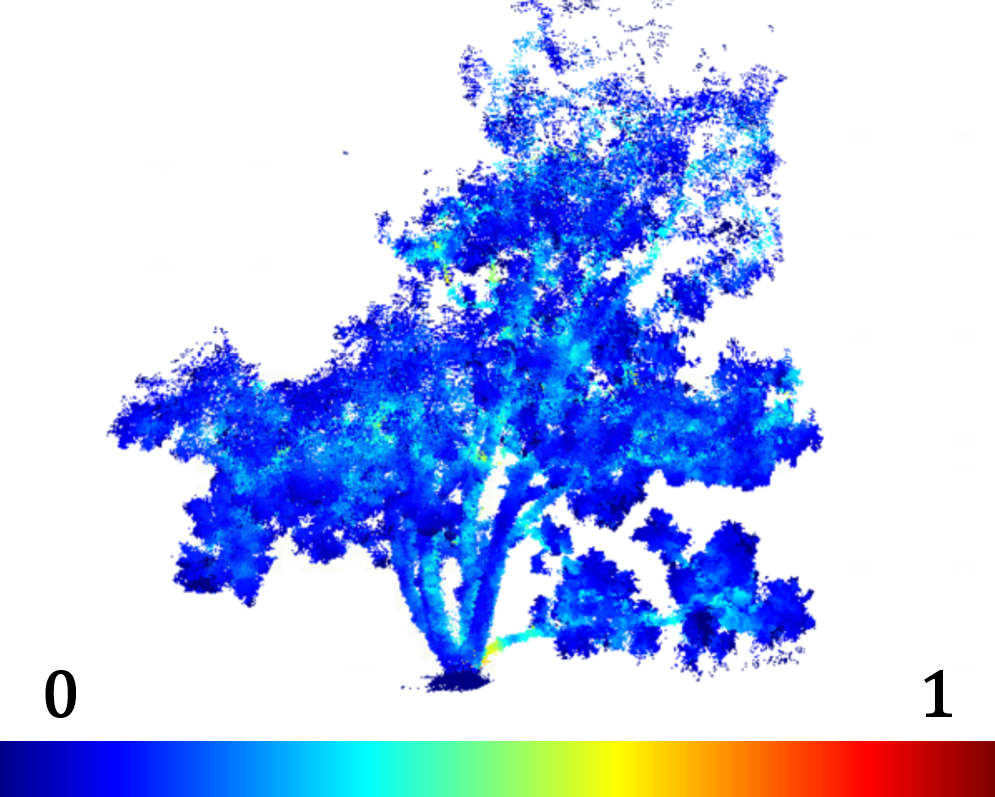}
        \caption{Raw graph score}
    \end{subfigure}
    ~
    \begin{subfigure}[t]{\columnwidth}
        \centering
        \includegraphics[width=\textwidth]{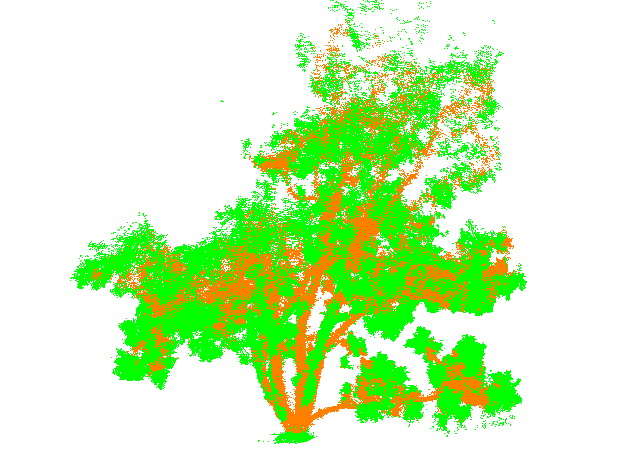}
        \caption{Classification following application of score threshold}
    \end{subfigure}
    \caption{Qualitative result of matter classification on a single real avocado tree.  (a) shows the raw graph score propagated from nodes to points, while (b) shows the result of classification through applying a percentile threshold, with orange points as woody matter and green points as leafy matter.}
    \label{fig:results-class-qual}
\end{figure}

Again we present a qualitative example of the matter classification operation in Figure~\ref{fig:results-class-qual}.
For these experiments we used the same dataset as in the ones for segmentation, though here we score results with the F1 score derived from the binary classification of leafy versus woody matter.

The average F1 score reported across all real data was 0.482, while virtual data reported an average F1 of 0.488.

\subsubsection{Virtual data}

\begin{figure}[ht]
    \centering
    \includegraphics[width=\columnwidth]{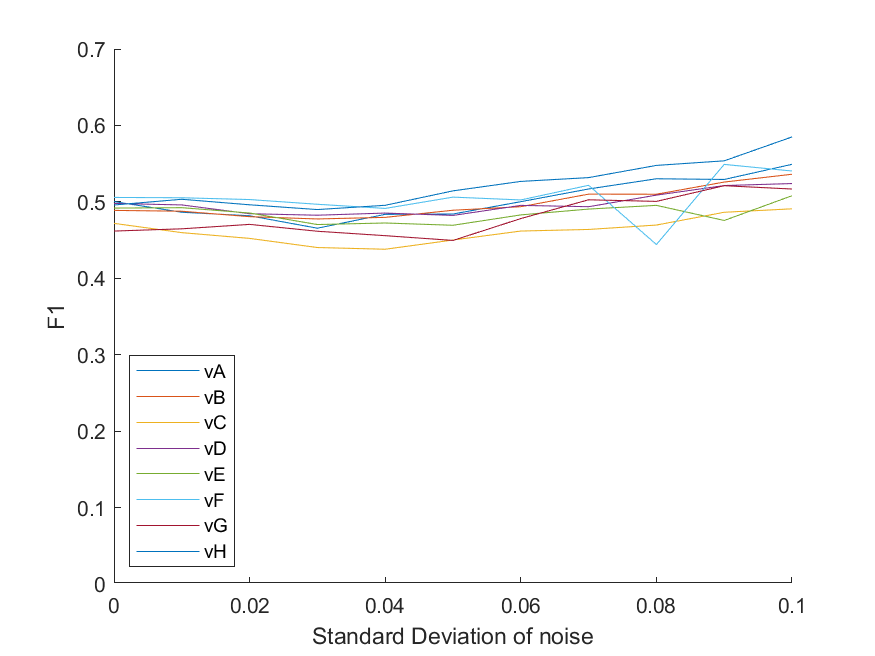}
    \caption{Results of matter classification on virtual data.  Data were generated using consistent tree spacing of 6m across various levels of introduced Gaussian noise.}
    \label{fig:results-class-virt}
\end{figure}

Figure~\ref{fig:results-class-virt} presents the results of 8 randomly generated stands of trees across 11 levels of introduced noise.

\subsubsection{Real data}

\begin{figure}[ht]
    \centering
    \includegraphics[width=\columnwidth]{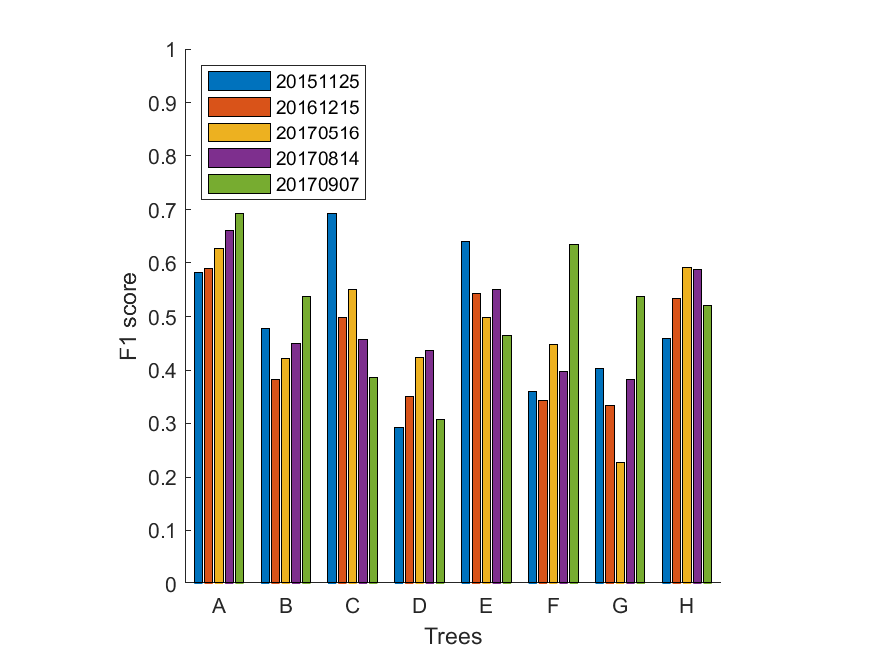}
    \caption{Results of our classification on real data.}
    \label{fig:results-class-real}
\end{figure}

Figure~\ref{fig:results-class-real} shows the quantitative results of applying matter classification on real avocado trees.
Again, we present results using both unweighted graphs and graphs weighted by cosine similarity of enriched features.

\begin{figure}[ht]
    \centering
    \begin{subfigure}[t]{\columnwidth}
        \centering
        \includegraphics[width=\textwidth]{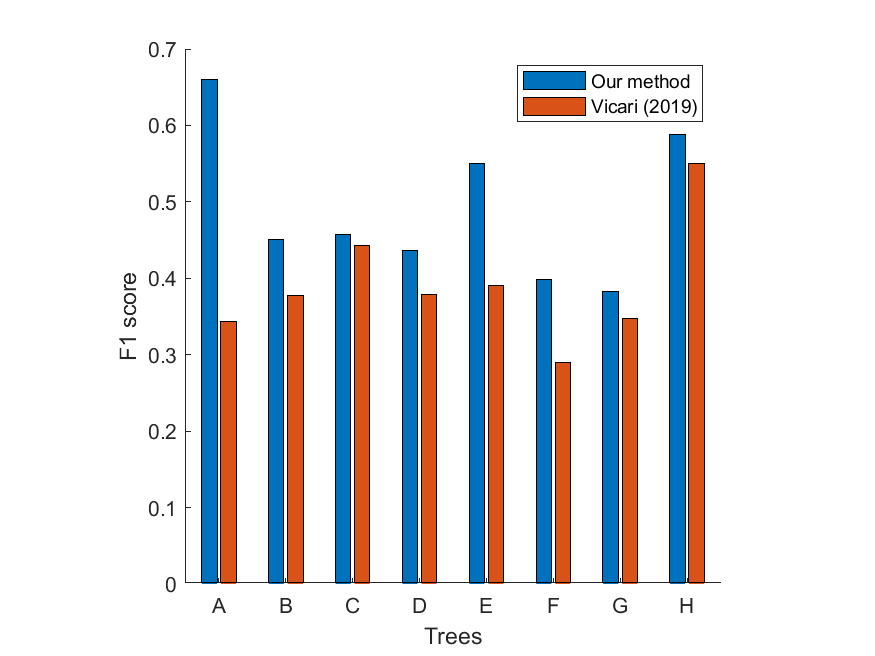}
        \caption{F1 score}
    \end{subfigure}
    ~
    \begin{subfigure}[t]{\columnwidth}
        \centering
        \includegraphics[width=\textwidth]{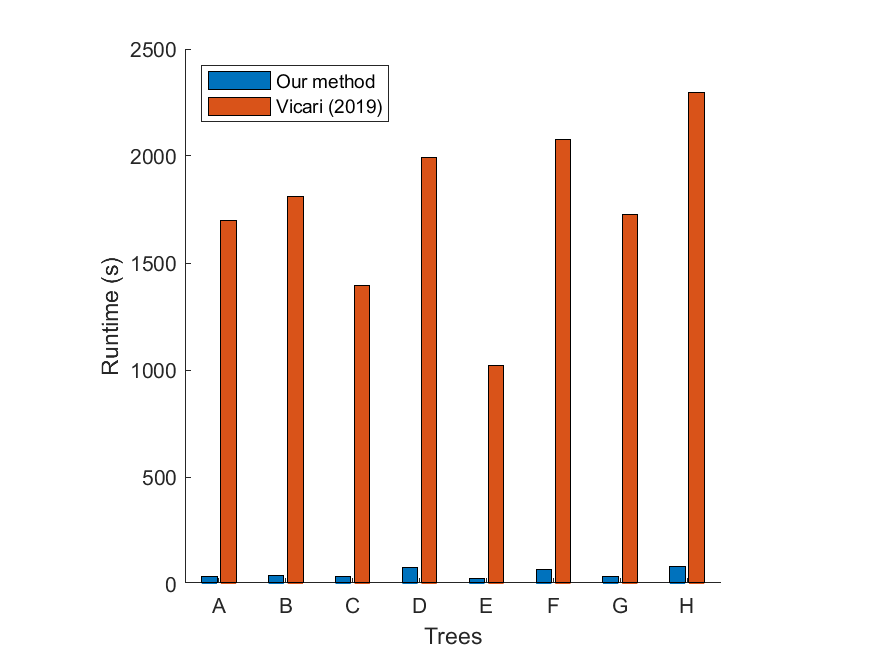}
        \caption{Runtime}
    \end{subfigure}
    \caption{Comparison between our method and that of \cite{vicari2019leaf} for real avocado tree scans.}
    \label{fig:results-class-vicari}
\end{figure}

Finally, we present results in Figure~\ref{fig:results-class-vicari} of a comparison on the results on our real data of our method and the method presented by ~\cite{vicari2019leaf}.
For these experiments, we ran both methods with default parameters (for the Vicari method, voxel size=0.05 and steps=40) on one capture set of avocado data (8 datasets representing 24 trees).
Experimentation with different parameters using Vicari's method was infeasible due to runtime.

\subsection{Enrichment}

Table~\ref{tab:enrich-results} presents the results for individual tree segmentation and matter classification, as the average v-measure or F1 score respectively, for real and virtual data.
Here we do not provide breakdowns of results for brevity as they did not present as significantly different on an individual scale.

\begin{table}[ht]
    \centering
    \begin{tabular}{lrr}
                                & Real  & Virtual \\
        Segmentation plain      & 0.915 & 0.884   \\
        Segmentation enriched   & 0.907 & 0.842   \\
        Classification plain    & 0.482 & 0.488   \\
        Classification enriched & 0.420 & 0.512
    \end{tabular}
    \caption{Aggregated results for segmentation and classification on enriched an unenriched (plain) data.  Plain data was computed using just XYZ coordinates of points, while enriched data was computed by taking graph edge weight as the cosine similarity between connected nodes using all enriched features listed in Table~\ref{tab:enrichment} }
    \label{tab:enrich-results}
\end{table}

\section{Discussion}
\label{sec:discussion}

\subsection{Trunk detection}
Table~\ref{tab:results-findtrunks} shows the results of finding trunks, with an F1 score of 0.774 on real data.
A likely cause for false negatives in real data are the high-density (2m spacing) trees which are visible in Figure~\ref{fig:results-ft-qual-hd}.
These trees are planted close together and, in the case of trellises, feature canopy close to the ground, which makes identification of a single trunk per tree difficult.
Furthermore, the mango data contains fence posts and non-tree items (e.g. vehicles) which would register false positives (for example the post on the far right of Figure~\ref{fig:results-ft-qual-hd}).
The average distance accuracy reported is 0.357m.  The trees in the real data have trunks with a diameter on the order of 0.2-0.5m, and the ground truth points are generated manually by selecting a scanned point on one side of the tree, so this accuracy is within expected range for correct detections.  For example, our method may in some cases choose a point on the other side of the trunk from the manual labelling.
Our method outperforms that of \cite{itakura2018automatic}.  The latter has more TPs but also many more FPs, suggesting that on our data the method generates too many candidate points.

In the virtual data, the method performs slightly worse with an F1 score of 0.687.
A possible reason for this can be seen in Figure~\ref{fig:results-ft-qual}, where trees with branches close to the ground are counted as having multiple trunks.
The method compensates for multiple trunk points detected within a specified radius, so where the tree spacing is known this radius can be calibrated to reduce this error (though this is out of scope for our experiments, as we intend these methods to be generalisable).
This effect is potentially exacerbated in Itakura's method, which reported a significantly higher FP on our virtual data despite a good recall of 0.859.

The response of F1 score against introduced Gaussian noise as presented in Figure~\ref{fig:results-findtrunks-noise} shows a slight decline in F1 score as noise increases, but it is certainly not significant.

\subsection{Individual segmentation}

In segmentation, the majority of points can be easily identified by simple methods like the closest-trunk approach, and this is reflected in the generally high scores in this area.  The average v-measures reported on our real data by the closest-trunk method, the \cite{kartal2020segmentation} method and our method were 0.826, 0.791 and 0.915 respectively.
However, our method qualitatively performs better in the overlapping sections shown in Figure~\ref{fig:results-seg-qual}, which represent a small proportion of points in the entire point cloud but can be significant, for instance if interested in identifying which branch belongs to which tree.
We explore this relationship in Figure~\ref{fig:results-seg-comp-real}, which shows the v-measure between our method and the others specifically only on points where both methods agree.
The results show our method mostly outperforming that of Kartal, but interestingly the closest-trunk method performs better for lower values of v-measure.
One reason for this could be that not all points in the cloud fall within our constructed graph; any points which are too far from the main canopy to be connected by a graph edge are labelled as unknown, and these points can occur for a variety of reasons including wind or internal occlusion.
These points should instead be classified by the closest trunk method to achieve better results.
The runtime findings shown in Figure~\ref{fig:results-seg-times-real} suggest that this combination of approaches could also demonstrate efficiency, as the closest-trunk method presents a longer runtime in all cases (though it could doubtless be implemented more optimally), while if this approach was only used a smaller subset of unknown points this would not be a concern.

Experimentation on virtual data show that this method is reasonably stable when put against varying noise, as presented in Figure~\ref{fig:results-seg-noise}.
This is to be expected, since the method does not rely on geometric features but rather the relative distance between points.

However, results mapped against tree spacing presented in Figure~\ref{fig:results-seg-spacing} show a clear degradation of results from an average v-measure of 0.994 at 8m spacing to 0.766 at 3m spacing.
Again, this effect is expected as the overlap between trees becomes more significant as spacing decreases.

\subsection{Matter classification}

Classification struggled to perform well across the board, with an average F1 score of 0.482 for real data and 0.488 for virtual data.
Figure~\ref{fig:results-class-real} presents results for all real data, broken down by tree and capture, demonstrating that in general some trees perform well (Tree A reports an F1 of 0.58 to 0.69), while others (e.g. Tree D) report F1 scores closer to 0.3.
Poor results may be due to varying levels of visibility and occlusion leading to difficulty in both automated classification and manual labelling, lower-quality point clouds due to wind during scanning.

As presented in Figure~\ref{fig:results-class-vicari}, our method outperforms the state-of-the-art approach presented by \cite{vicari2019leaf} by an average F1 score of 0.1 (for the capture dataset presented, our method displayed an average F1 of 0.490 against Vicari's average F1 of 0.390).
Part of the reason for this could be that Vicari's method uses the geometric features of the point cloud as well as the structural characteristics, and this approach is made more challenging by the noise and variation in scan quality inherent to captures by mobile LiDAR.
However, Figure~\ref{fig:results-class-vicari} also presents the runtimes of the two methods on the same data, in which our method significantly it, with an average runtime for these trees of 49s against 1752s.

\begin{figure}[ht]
    \centering
    \begin{subfigure}[t]{0.24\textwidth}
        \centering
        \includegraphics[width=\textwidth]{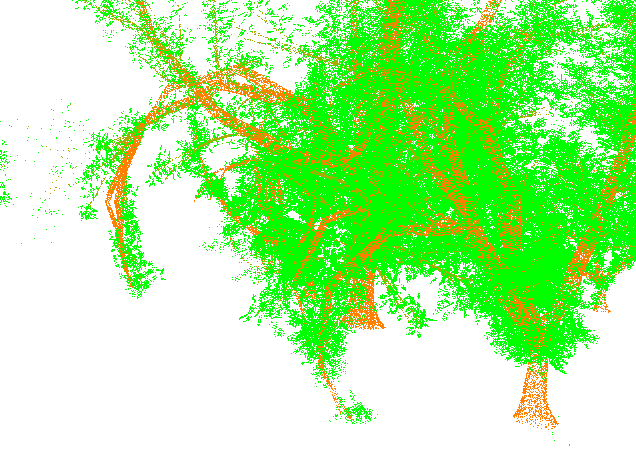}
        \caption{$\sigma = 0$}
    \end{subfigure}%
    ~
    \begin{subfigure}[t]{0.24\textwidth}
        \centering
        \includegraphics[width=\textwidth]{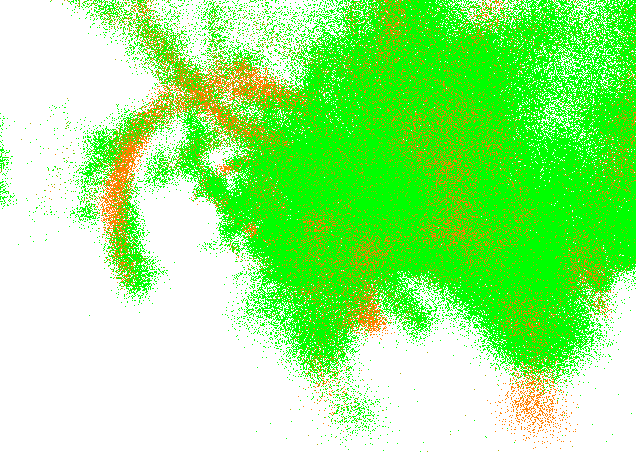}
        \caption{$\sigma = 0.1$}
    \end{subfigure}
    \caption{Virtual trees generated without noise and large noise.}
    \label{fig:disc-noise}
\end{figure}

As with segmentation, Figure~\ref{fig:results-class-virt} show our classification is almost unaffected by noise levels whose range is demonstrated in Figure~\ref{fig:disc-noise}.
Our method is not based on identification of geometric features, so despite the destruction of detail seen at $\sigma = 0.1$, the tree structure should remain identifiable.
In fact the average F1 score increases slightly with the noise (from 0.489 at $\sigma=0$ to 0.531 at $\sigma=0.1$), though the standard deviation almost doubles from 0.015 to 0.029.  The reason for this relationship is hard to discern, though may be due to the voxel propagation at the conclusion of classification.  As the noise increases, individual points will move in and out of voxels identified as trunk matter, improving or worsening detections at random.

One reason the classification results are poor could be that due to sensor noise and occlusion, these labels are imperfect.
Higher in the canopy, the branches tend to be thinner and the distance from the sensor is greater, leading to less distinction between leaves and branches.
This problem is also exacerbated by wind causing movement of elements between successive LiDAR passes.

Another issue with our method is that the graph search does not pass through every node, instead finding central "highways" to travel through.
In areas where the voxel size is approximately the same as the trunk thickness, most trunk points are correctly identified, but where the trunk is multiple nodes wide, many nodes are bypassed and therefore likely to be classified as leaves.

Further study is required to understand the pros and cons of our method.
More real data would be beneficial to this understanding, using a greater variety of scanners.
Our study was focused on handheld LiDAR, but application to tripod, aerial and mobile LiDAR may demonstrate versatility to point cloud noise and density which we currently only theorise using our artificial noise experiments.

\subsection{Enrichment}

The point enrichment scheme shows promise when examined qualitatively (as in Figure~\ref{fig:method-graphop-minigraph}), but displays little or no improvement on final results.
This could be partly due to the large number of introduced features - an ablation study could clarify which ones are significant and discarding the rest may improve results.
The enriched features could also be used more effectively.  Further study could show positive results in applying machine learning to the enriched features to improve the initial results provided by the graph operation.

Taking the method as it stands, the ability to simply understand the structure of the trees and the connectivity of the branches has further applications.
In particular, we are interested in studying the effects of pruning on the canopy, which can be simulated using the knowledge of tree connectivity.

\section{Conclusions}
\label{sec:conclusion}
We presented a system for processing point cloud data captured using LiDAR at a fruit orchard to detect trunk location with no priors, segment individual trees even in high-density contexts, and classify trunk and leaf matter automatically.
Detecting trunk locations displayed an average F1 score of 0.774, a 0.237 improvement over a similar method in the literature, as well as a smaller distance accuracy for true positive detections.
Individual tree segmentation demonstrated an average v-measure result of 0.915 on real data, and performed qualitatively well on overlapping trees, though this was less clear quantitatively.
Matter classification outperformed an existing method by an F1 margin of 0.1 with significantly lower runtimes, but only achieved an average F1 score of 0.490 on real data.
Experimental results using virtual data suggested that our methods were stable with respect to point cloud noise but suffered when tree spacing was reduced.

\subsubsection*{Acknowledgements}
This work is supported by the Australian Centre for Field Robotics (ACFR) at The University of Sydney. Thanks to Vsevolod (Seva) Vlaskine and the software team for their continuous support, and to Salah Sukkerieh for his continuous direction and support of Agriculture Robotics at the ACFR. For more information about robots and systems for agriculture at the ACFR, please visit http://sydney.edu.au/acfr/agriculture.

\bibliography{references}{}

\end{document}